\documentclass[10pt,twocolumn,letterpaper]{article}

\usepackage{iccv}
\usepackage{times}
\usepackage{epsfig}
\usepackage{graphicx}
\usepackage{amsmath}
\usepackage{amssymb}

\usepackage{mathtools,xparse}
\usepackage{booktabs}
\usepackage{moresize}
\usepackage{arydshln}

\DeclarePairedDelimiter{\norm}{\lVert}{\rVert}

\usepackage[pagebackref=true,breaklinks=true,letterpaper=true,colorlinks,bookmarks=false]{hyperref}

\usepackage[breaklinks=true,bookmarks=false]{hyperref}

\iccvfinalcopy 

\def\iccvPaperID{****} 

\begin{document}

\title{Generating Multiple Diverse Hypotheses for Human 3D Pose Consistent with 2D Joint Detections}

\author{Ehsan Jahangiri, Alan L. Yuille\\
Johns Hopkins University, Baltimore, USA\\
{\tt\small ejahang1@jhu.edu, alan.yuille@jhu.edu}
}


\maketitle
\vspace{-5mm}
\begin{abstract}
We propose a method to generate multiple diverse and valid human pose hypotheses in 3D all consistent with the 2D detection of joints in a monocular RGB image. We use a novel generative model uniform (unbiased) in the space of anatomically plausible 3D poses. Our model is compositional (produces a pose by combining parts) and since it is restricted only by anatomical constraints it can generalize to every plausible human 3D pose. Removing the model bias intrinsically helps to generate more diverse 3D pose hypotheses. We argue that generating multiple pose hypotheses is more reasonable than generating only a single 3D pose based on the 2D joint detection given the depth ambiguity and the uncertainty due to occlusion and imperfect 2D joint detection. We hope that the idea of generating multiple consistent pose hypotheses can give rise to a new line of future work that has not received much attention in the literature. We used the Human3.6M dataset for empirical evaluation.
\end{abstract}





\section{Introduction}

\label{sec:intro}

\begin{figure}[t]
\centering
\vspace{3mm}
\includegraphics[width=0.45\textwidth]
{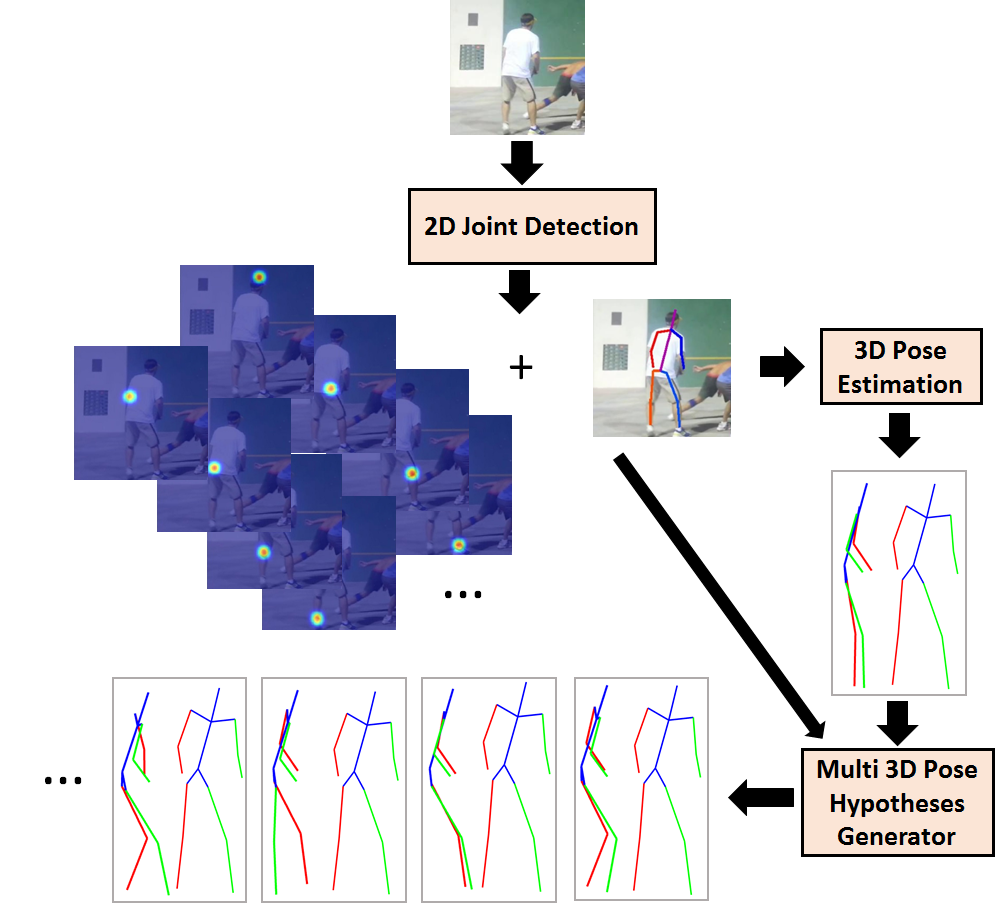}
\vspace{2mm}
\caption[Overview]{\footnotesize{The input monocular image is first passed through a CNN-based 2D joint detector which outputs a set of heatmaps for soft localization of 2D joints. The 2D detections are then passed to a 2D-to-3D pose estimator to obtain an estimate of the 3D torso and the projection matrix. Using the estimated 3D torso, the projection matrix, and the output of the 2D detector we generate multiple diverse 3D pose hypotheses consistent with the output of 2D joint detector.}}
\label{fig:overview}
\vspace{-4mm}
\end{figure}

Estimating the 3D pose configurations of complex articulated objects such as humans from monocular RGB images is a challenging problem. There are multiple factors contributing to the difficulty of this critical problem in computer vision: (1) multiple 3D poses can have similar 2D projections. This renders 3D human pose reconstruction from its projected 2D joints an ill-posed problem; (2) the human motion and pose space is highly nonlinear which makes pose modeling difficult; (3) detecting precise location of 2D joints is challenging due to the variation in pose and appearance, occlusion, and cluttered background. Also, minor errors in the detection of 2D joints can have a large effect on the reconstructed 3D pose. These factors favor a 3D pose estimation system that takes into account the uncertainties and suggests multiple possible 3D poses constrained only by reliable evidence. Often in the image, there exist much more detailed information about the 3D pose of a human than the 2D location of the joints (such as contextual information and difference in shading/texture due to depth disparity). Hence, most of the possible 3D poses consistent with the 2D joint locations can be rejected based on more detailed image information (\eg in an analysis-by-synthesis framework or by investigating the image with some mid-level queries such as ``Is the left hand in front of torso?'') or by physical laws (\eg gravity). We can also imagine scenarios where the image does not contain enough information to rule out or favor one 3D pose configuration over another especially in the presence of occlusion. In this paper, we focus on generating multiple plausible and diverse 3D pose hypotheses which while satisfying humans anatomical constraints are still consistent with the output of the 2D joint detector. Figure~\ref{fig:overview} illustrates an overview of our approach. 

\begin{figure}[t]
\centering
\includegraphics[height = 0.225\textwidth, width=0.225\textwidth]{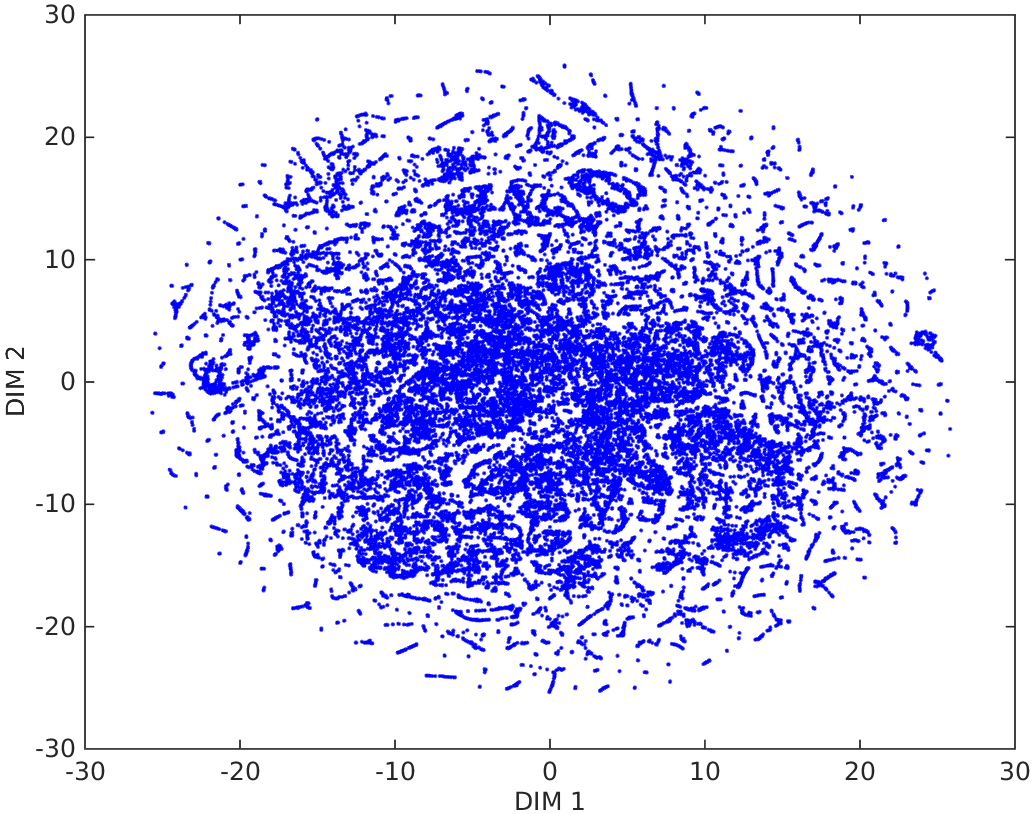}
\includegraphics[height = 0.225\textwidth, width=0.225\textwidth]{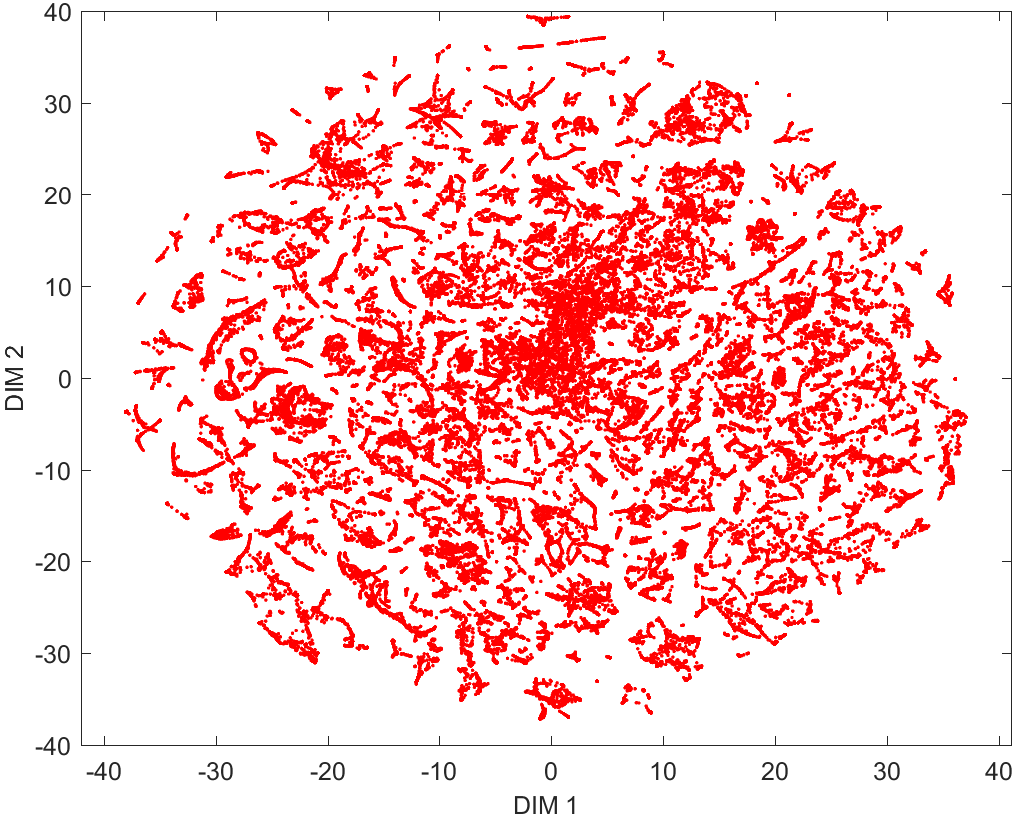} \\
(a) \\
\includegraphics[width=0.11\textwidth]{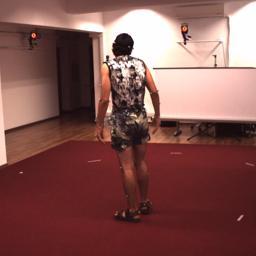}
\includegraphics[width=0.11\textwidth]{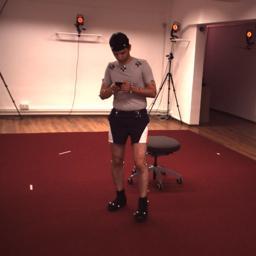}
\includegraphics[width=0.11\textwidth]{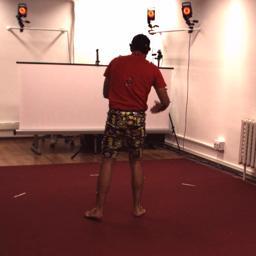}
\includegraphics[width=0.11\textwidth]{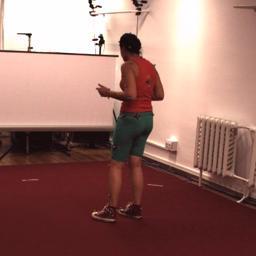}\\
\includegraphics[width=0.11\textwidth]{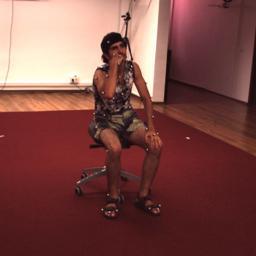}
\includegraphics[width=0.11\textwidth]{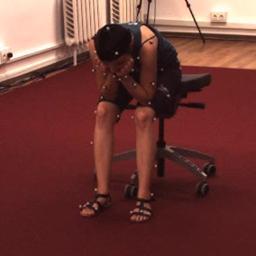}
\includegraphics[width=0.11\textwidth]{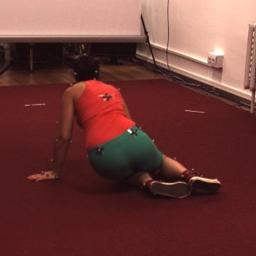}
\includegraphics[width=0.11\textwidth]{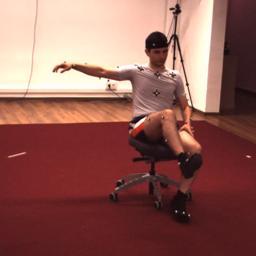}
(b)
\caption[tSNE]{\footnotesize{(a): The t-SNE visualization of poses from the H36M (fist from left) and ABCD (second from left). (b): The images corresponding to the random selection of poses from the dense (top row in right) and sparse (bottom row in right) area of the H36M t-SNE map confirm the dataset bias toward standing poses compared to sitting poses.}}
\label{fig:tSNE}
\vspace{-5mm}
\end{figure}

The space of valid human poses is a non-convex complicated space constrained by the anatomical and anthropomorphic limits. A bone never bends beyond certain angles with respect to its parent bone in the kinematic chain and its normalized length, with respect to other bones, cannot be much shorter/longer than standard values. This inspired Akhter and Black~\cite{Akhter2015} to build a motion capture dataset composed of 3D poses of flexible subjects such as gymnasts and martial artists to study the joint angle limits. The statistics of 3D poses in this motion capture dataset is different from the previously existing motion capture datasets such as CMU~\cite{CMUDataset}, Human 3.6M~\cite{h36m_pami}, and HumanEva~\cite{HumanEva10}, because of their intention to explore the joint angle limits rather than performing and recognizing typical human actions. Figure~\ref{fig:tSNE} shows the t-SNE visualization~\cite{tSNE_2008} of poses from {\bf A}khter\&{\bf B}lack motion {\bf C}apture {\bf D}ataset ({\bf ABCD}) versus H36M in two dimensions. One can see that the ``ABCD'' dataset is more uniformly distributed compared to the H36M dataset. We randomly selected 4 poses from the dense and surrounding sparse areas in the H36M t-SNE map and have shown the corresponding images. One can see that all of the four samples selected from the dense areas correspond to standing poses whereas all of the four samples selected from sparse areas correspond to sitting poses. 

Training and testing a 3D model on a similarly biased dataset with excessive repetition of some poses will result in reduced performance on novel or rarely seen poses. As a simple demonstration, we learned a GMM 3D pose model~\cite{GMM_2004} from a uniformly sampled set of Human 3.6M poses (all 15 actions) and evaluated the likelihood of 3D poses per action under this model. The average likelihood per action (up to a scaling factor) was: Directions 0.63, Discussion 0.74, Eating 0.56 , Greeting 0.63 , Phoning 0.28 , Posing 0.38 , Purchases 0.55 , Sitting 0.07 , Sitting Down 0.07 , Smoking 0.47 , Taking Photo 0.23 , Waiting 0.33 , Walking 0.64 , Walking Dog 0.29 , and Walk Together 0.25. According to the GMM model, the ``Discussion'' poses are on average almost 10 times more likely than ``Sitting'' poses which is due to the dataset and consequently the model bias. The EM algorithm used to learn the GMM model attempts to maximize the likelihood of all samples which will lead to a biased model if the training dataset is biased. Obviously, any solely data-driven model learned from a biased dataset that does not cover the full range of motion of human body can suffer from lack of generalization to novel or rarely seen yet anatomically plausible poses.

We propose a novel generative model on human 3D poses uniform in the space of physically valid poses (satisfying the constraints from~\cite{Akhter2015}). Since our model is constrained only by the anatomical limits of human body it does not suffer from dataset bias which is intrinsically helpful to diversify pose hypotheses. Note that the pose-conditioned anatomical constraints calculated in~\cite{Akhter2015} was originally used in a constrained optimization framework for single 3D pose estimation and turning those constraints into a generative model to produce uniform samples is not trivial.  One of our main contributions is a pose-conditioned generative model which has not been done previously. We generate multiple anatomically-valid and diverse pose hypotheses consistent with the 2D joint detections to investigate the importance of having multiple pose hypotheses under depth and missing-joints (\eg caused by occlusion) ambiguities. In the recent years, we have witnessed impressive progress in accurate 2D pose estimation of human in various pose and appearances which is made possible thanks to deep neural networks and lots of annotated 2D images. We take advantage of the recent advancement in human 2D pose estimation and seed our multi-hypotheses pose generator by an off-the-shelf 3D pose estimator. Namely, we use the ``Stacked Hourglass'' 2D joint detector~\cite{Newell_2016} and the 2D-to-3D pose estimators of Akhter\&Black~\cite{Akhter2015} and Zhou et al.~\cite{ZhouUpenn2015} to estimate the 3D torso and projection matrix. However, note that to our generic approach does not rely on any specific 2D/3D pose estimator and can easily adopt various 2D/3D pose estimators. 

After briefly discussing some related works in subsection~\ref{Sec:Related_Work} we propose our approach in section~\ref{Sec:Proposed_Method}. Our experimental results based on multiple 3D pose estimation baselines is given in section~\ref{Sec:Results}. We conclude in section~\ref{Sec:Conclusion}.

\subsection{Related Work}
\label{Sec:Related_Work}
There are quite a few works in the human pose estimation literature that are directly or indirectly related to our work. Reviewing the entire literature is obviously beyond the scope of this paper. Several areas of research are related to our work such as 2D joiont detection, 3D pose estimation, and generative 3D pose modeling. Due to the advancements made by deep neural networks, the most recent works on 2D joint detection are based on convolutional neural networks (CNN)~\cite{toshev2014deeppose,chen2014articulated,tompson2014joint,chu2016structured,yang2016end,chu2016crf,Wei_2016,Newell_2016,bulat2016human,Rogez2017} compared to the traditional hand-crafted feature based methods~\cite{Sapp2013,Yang2011,Eichner12}. On the other hand, most of the 3D pose estimation methods use sparse coding based on an overcomplete dictionary of basis poses to represent a 3D pose and fit the 3D pose projection to the 2D joint detections~\cite{Ramakrishna_2012,Chunyu_2014,Akhter2015,ZhouUpenn2015,Zhou_2016}. Some works~\cite{Deep3DPose,rogez2016mocap,Rogez2017} try to train a deep network to directly predict 3D poses. However, purely discriminative approaches for 3D structure prediction (such as ~\cite{Deep3DPose}) are usually very sensitive to data manipulation. On the other hand, it has been shown that the deep networks are very effective and more robust at detecting 2D templates (compared to 3D structures) such as human 2D body parts in images~\cite{Newell_2016}.

We use conditional sampling from our generative model to generate multiple consistent pose hypotheses. A number of previous works~\cite{BureniusSC13,Sigal_IJCV_11,amin13bmvc,belagiannis20143d,belagiannis20163Dpami} have used sampling for human pose estimation. However, the sampling performed by these works are for purposes different from our goal to generate multiple diverse and valid pose hypotheses. For example, Amin et al.~\cite{amin13bmvc} use a mixture of pictorial structures and perform inference in two stages where the first stage reduces the search space for the second inference stage by generating samples for the 2D location of each part. 

Some more closely related works include~\cite{Sminchisescu2003, Moll2011, Lee2004, Park2011, PonsMoll_CVPR2014, Simo_cvpr2013, Lehrmann2013,SimoSerraCVPR2012}. Sminchisescu and Triggs~\cite{Sminchisescu2003} search for multiple local minima of their fitting cost function using a sampling mechanism based on forwards/backwards link flipping to generate pose candidates. Pons-Moll et al.~\cite{Moll2011} use inverse kinematics to sample the pose manifold restricted by the input video and IMU sensor cues in a particle filter framework. Lee and Cohen~\cite{Lee2004} use proposal maps to consolidate the evidence and generating 3D pose candidates during the MCMC search where they model the measurement uncertainty of 2D position of joints using a Gaussian distribution. Their MCMC approach suffers from high computational cost. Park and Ramanan~\cite{Park2011} generate non-overlapping diverse pose hypotheses (only in 2D) from a part model. One interesting work is the ``Posebit'' by Pons-Moll et al.~\cite{PonsMoll_CVPR2014} that can retrieve pose candidates from a MoCap dataset of 3D poses given answers to some mid-level queries such as ``Is the right hand in front of torso?'' using decision trees. This approach is heavily dependent on the choice of MoCap dataset and cannot generalize to unseen poses. Simo-Serra1 et al.~\cite{Simo_cvpr2013} model the 2D and 3D poses jointly in a Bayesian framework by integrating a generative model and discriminative 2D part detectors based on HOGs. Lehrmann et al.~\cite{Lehrmann2013} learn a generative model from the H36M MoCap dataset whose graph structure (not a Kinematic chain) is learned using the Chow-Liu algorithm. Simo-Serra et al.~\cite{SimoSerraCVPR2012} propagate the error in the estimation of 2D joint locations (modeled using Gaussian distributions) into the weights of dictionary elements in a sparse coding framework; then by sampling the weights, some 3D pose samples are generated and sorted based on the SVM score on joint distance features. However, their approach does not guarantee that the joint angle constraints are satisfied and do not address the depth ambiguity. We impose ``pose-conditioned'' joint angle and bone length constrains to ensure pose validity of samples from our generative model which has not been done before. In addition, our unbiased generative model restricted only by anatomical constrains helps in generating more diverse 3D pose hypotheses.

\section{The Proposed Method}
\label{Sec:Proposed_Method}
\begin{figure}[t]
\centering
\includegraphics[width=0.45\textwidth]{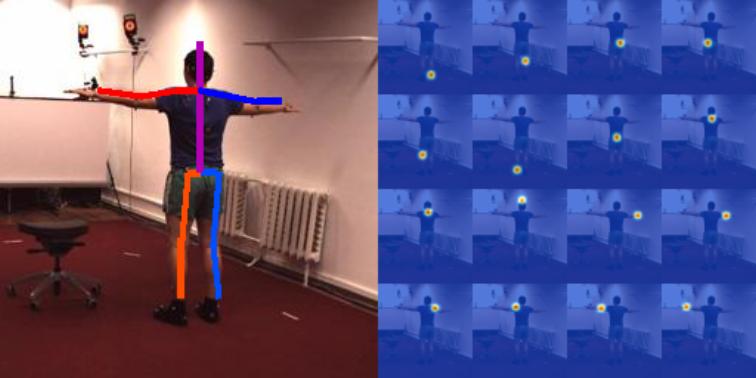} 
\includegraphics[width=0.45\textwidth]{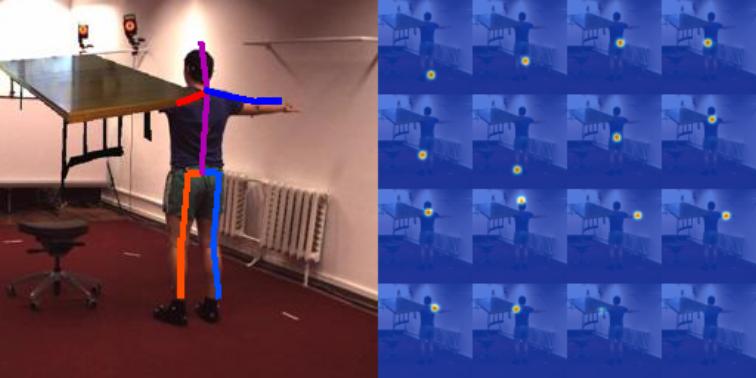}
\caption[Hourglass]{\small{``Stacked Hourglass'' 2D joint detector~\cite{Newell_2016} in the absence and presence of occlusion. On the right-hand-side of each image are the corresponding heatmaps for joints.}}
\label{fig:HourglassOcc}
\vspace{-3mm}
\end{figure}
Since our approach is closely related to the joint-angle constraints used in~\cite{Akhter2015}, we find it helpful for better readability to briefly review this work. To represent the human 3D pose by its joints let $\mathbf{X}$ denote the matrix corresponding to $P$ kinematic joints in the 3D space namely $\mathbf{X} = [\mathbf{X}_1 ... \mathbf{X}_P] \in \mathcal{X} \subset {\rm I\!R}^{3\times P}$ where $\mathcal{X}$ denotes the space of valid human poses. Akhter\&Black~\cite{Akhter2015} (similar to~\cite{Ramakrishna_2012,ZhouUpenn2015}) assumed that all of the 2D joints are observed and estimated a single 3D pose by solving the following optimization problem:
\vspace{-2mm}
\begin{align}
\label{AkhterOpt}
\min_{\mathbf{\omega},s,\mathbf{R}} \hspace{0.5mm} C_r + C_p + \beta C_l,
\vspace{-2mm}
\end{align}
where, $C_r$ is a measure of fitness between the estimated 2D joints $\hat{\mathbf{x}} \in {\rm I\!R}^{2\times P}$ and the projection and translation of estimated 3D pose $\hat{\mathbf{X}} = [\hat{\mathbf{X}}_1 ... \hat{\mathbf{X}}_P] \in {\rm I\!R}^{3\times P}$ to the 2D image coordinate system in a weak perspective camera model (orthographic projection) with scaling factor $s \in {\rm I\!R}^{+}$, rotation $\mathbf{R} \in SO(3)$, and translation $\mathbf{t} \in {\rm I\!R}^{2\times 1}$, defined as:
\vspace{-2mm}
\begin{align}
\label{Cr}
C_r = \sum_{i=1}^{P} \norm{\hat{\mathbf{x}}_i - s \mathbf{R}_{1:2} \hspace{0.5mm} \hat{\mathbf{X}}_i + \mathbf{t}}_{2}^2,
\vspace{-2mm}
\end{align}
where, $\mathbf{R}_{1:2}$ denotes the first two rows of the rotation matrix. Note that if the origin of the 3D world coordinate system gets mapped to the origin of the 2D image coordinate system then $\mathbf{t} = \mathbf{0}$; this is usually implemented by centering the 2D and 3D poses. Authors used a sparse representation of the 3D poses similar to~\cite{Ramakrishna_2012} where the 3D pose is represented by a sparse linear combination of bases selected using the Orthogonal Matching Pursuit (OMP) algorithm~\cite{Mallat93} from an overcomplete dictionary of pose atoms, namely $\hat{\mathbf{X}} = \mathbf{\mu} + \sum_{i \in \mathcal{I^*}} \omega_i \mathbf{D}_i$, where $\mathbf{\mu}$ is the mean pose obtained by averaging poses from the CMU motion capture dataset~\cite{CMUDataset} and $\mathcal{I^*}$ denotes the indices of selected bases using OMP with weights $\mathbf{\omega}_i$. An overcomplete dictionary of bases was built by concatenating PCA bases from poses of different action classes in the CMU dataset after bone length normalization and Procrustes aligned. The second term $C_p$ in equation~\eqref{AkhterOpt} is equal to zero if the estimated pose $\hat{\mathbf{X}}$ has valid joint angles for limbs and infinity otherwise. According to the pose-conditioned constraints in~\cite{Akhter2015} a pose has valid joint angles if the upper arms/legs' joint angles map to a 1 in the corresponding occupancy matrix (learned from the ABCD dataset) and the lower arms/legs satisfy two conditions that prevent these bones from bending beyond feasible joint-angle limits (inequalities~\eqref{LowerArmLegConst} and~\eqref{BoundingBoxConst}). The term $C_l$ in equation~\eqref{AkhterOpt} penalizes the difference between the squares of the estimated $i^{\text{th}}$ bone length $l_i$ and the normalized mean bone length $\bar{l}_i$ \ie, $C_l = \sum_{i=1}^{N} |l_i^2 - \bar{l}_i^2|$ (normalized mean bones calculated from the CMU dataset) with weight $\beta$. Note that~\cite{Akhter2015} does not introduce any generative pose model.

As we mentioned earlier, 3D pose estimation from 2D landmark points in monocular RGB images is inherently an ill-posed problem because of losing the depth information. There can be multiple valid 3D poses with similar 2D projection even if all of the 2D joints are observed (see Figure~\ref{fig:overview}). The uncertainty and number of possible valid poses can further increase if some of the joints are missing. The missing joints scenario is more realistic because it happens when either these joints exist in the image but are not confidently detected, due to occlusion and clutter, or do not exist within the borders of the image \eg when only the upper body is visible similar to images from the FLIC dataset~\cite{Sapp2013}. It is observed that thresholding the confidence score obtained from some deep 2D joint detectors (\eg ~\cite{Newell_2016,pishchulin16cvpr,insafutdinov16ariv}) can be reasonably used as an indicator for the confident detection of a joint. Figure~\ref{fig:HourglassOcc} shows the the output of ``Stacked Hourglass'' 2D joint detector~\cite{Newell_2016} in the absence and presence of a table occluder segmented out from the Pascal VOC dataset~\cite{Everingham15} and pasted on the left hand of the human subject. On the right-hand-side of each image is shown the heatmap for each joint. It can be seen that the level of the two heatmaps corresponding to the left elbow and left wrist drop after placing the table occluder on the left hand. Newell et al.~\cite{Newell_2016} used the heatmap mean as a confidence measure for detection and threshold it at 0.002 to determine visibility of a joint. Obviously, invisibility of some joints in the image can result in multiple hallucinations for the 2D/3D locations of the joints. Let $S_{\text{o}}$ and $S_{\text{m}}$ denote the set of observed and missing joints, respectively. We have $S_{\text{o}} \cap S_{\text{m}} = \emptyset$ and $S_{\text{o}} \cup S_{\text{m}} = \{1,2,...,P\}$, and let $\mathbf{\alpha} = \{\alpha_i \}_{i \in S_{\text{o}}}$ denote a set of normalized joint scores from the 2D joint detectors such that $\frac{1}{|S_{\text{o}}|} \sum_{i \in S_{\text{o}}} \alpha_i = 1$. The missing joints are detected by comparing the confidence score of 2D joint detector with a threshold (0.002 in the case of using Hourglass). For the case of missing joints, we modify the fitness measure to:
\vspace{-1mm}
\begin{align}
\label{Cr_missing}
C_r = \sum_{i \in S_{\text{o}}} \alpha_i \norm{\hat{\mathbf{x}}_i - s \mathbf{R}_{1:2} \hspace{0.5mm} \hat{\mathbf{X}}_i + \mathbf{t}}_{2}^2.
\vspace{-2mm}
\end{align}
The scores are normalized because they have to be in a comparable range with respect to the $C_l$ term in equation~\eqref{AkhterOpt} otherwise either $C_r$ is suppressed/ignored in the case of very small confidence scores or the same happens to $C_l$ in the case of very large scores. For example, if the mean of heatmaps from the Hourglass joint detector are directly (without normalization) used as scores the $C_r$ term will be drastically suppressed since the heatmaps are full of close-to-zero values. Note that the optimization problem in equation~\eqref{AkhterOpt} with the updated $C_r$ term according to equation~\eqref{Cr_missing} still outputs a full 3D pose even under missing joints scenario because the 3D pose is constructed by a linear combination of full body basis. However, there is no reason that the output 3D pose should have a close to correct 2D projection due to the missing joint ambiguity added to the depth ambiguity. Optimizing $C_r$ is a non-convex optimization problem over the 3D pose and projection matrix. To obtain an estimate of the 3D torso and projection matrix, we tried both iterating between optimizing over the projection matrix and 3D pose used in~\cite{Akhter2015} as well as the convex relaxation method in~\cite{ZhouUpenn2015} as will be presented in the experimental results section. Note that the torso pose variations are much fewer than the full-body. The torso plane is usually vertical and not as flexible as the full body. Hence, it is much easier to robustly estimate its 3D pose and the corresponding camera parameters. 

To generate multiple diverse 3D pose hypotheses consistent with the output of 2D joint detector, we cluster samples from a conditional distribution given the collected 2D evidence. For this purpose, we follow a rejection sampling strategy. Before discussing conditional sampling in subsection~\ref{Subsec:CondSamp} we describe unconditional sampling as follows.


\subsection{Unconditional Sampling}
\label{Subsec:UncondSamp}
Given the rigidity of human torso compared to the limbs (hands/legs), the joints corresponding to the torso including thorax, left/right hips, and left/right shoulders can be represented using a small size dictionary after an affine transformation/normalization. Given the torso, the upper arms/legs and head are anatomically restricted to be within certain angular limits. The plausible angular regions for the upper arms/legs and head can be represented using an occupancy matrix~\cite{Akhter2015}. This occupancy matrix is a binary matrix that assigns 1 to a discretized azimuthal $\theta$ and polar $\phi$ angle if these angles are anatomically plausible and 0 otherwise. These angular positions are calculated in the local Cartesian coordinate system whose two axis are the ``backbone'' vector and either the ``right shoulder $\rightarrow$ left shoulder'' vector (for the upper arms and head) or the ``right hip $\rightarrow$ left hip'' vector (for the upper hips). Hence, to generate samples for the upper arms/legs and head we just need to take samples from the occupancy matrix at places where the value is 1 and get the corresponding azimuthal and polar angles. Given the azimuthal and polar angles of the head we just need to travel in this direction for the length of the head; we do the same for the length of upper arms and legs to reach the elbows and knees, respectively. The normalized length of the bones is sampled from a Beta distribution with limited range under the constraint that similar bones have similar length \eg both upper arms have the same length.

According to~\cite{Akhter2015}, the lower arm/leg bone $\mathbf{b}_{p_1 \rightarrow p_2} = \mathbf{X}_{p_2} - \mathbf{X}_{p_1}$, where $p_2$ and $p_1$ respectively correspond to either ``wrist and elbow'' or ``ankle and knee'' is at a plausible angle if it satisfies two constraints. The first constraint is:
\vspace{-3mm}
\begin{align}
\label{LowerArmLegConst}
\mathbf{b}^\top \mathbf{n} + d < 0,
\end{align}
where $\mathbf{n}$ and $d$ are functions of the azimuthal $\theta$ and polar $\phi$ angles of their parent bone namely the upper arm or leg (resulting in pose-dependent joint angle limits) learned from the ABCD dataset. The above inequality defines a separating plane, with normal vector $\mathbf{n}$ and distance from origin $d$, that attempts to prevent the wrist and ankle from bending in a direction that is anatomically impossible. Obviously, for a very negative offset vector $d$ this constrain is always satisfied. Therefore, during learning of $\mathbf{n}$ and $d$ the second norm of $d$ is minimized, namely $\min_{\mathbf{n},d} \hspace{0.5mm} \norm{d}_2 \hspace{2mm} \text{s.t.} \hspace{2mm} \mathbf{B}^\top \mathbf{n} < -d \mathbf{1}$, where $\mathbf{B}$ is a matrix built by column-wise concatenation of all $\mathbf{b}$ instances in the ABCD dataset whose parents are at the same $\theta$ and $\phi$ angular location. 
The second constraint to satisfy is that the projection of normalized $\mathbf{b}$ (to unit length) onto the separating plane using the orthonormal projection matrix $\mathbf{T} = [\mathbf{T}_1; \mathbf{T}_2; \mathbf{T}_3]$, whose first row $\mathbf{T}_1$ is along $\mathbf{n}$, has to fall inside a bounding box with bounds $[bnd_1,bnd_2]$ and $[bnd_3,bnd_4]$, namely:
\begin{align}
\label{BoundingBoxConst}
&bnd_1 \le \mathbf{T}_2 \mathbf{b}/\norm{\mathbf{b}}_2 \le bnd_2,  \nonumber \\
&bnd_3 \le \mathbf{T}_3 \mathbf{b}/\norm{\mathbf{b}}_2 \le bnd_4,
\end{align}
where, bounds $bnd_1$, $bnd_2$, $bnd_3$, and $bnd_4$ are also learned from the ABCD dataset. To generate a sample $\mathbf{b}$ that satisfies the above constraints, we first generate two random values $u_2 \in [bnd_1,bnd_2]$ and $u_3 \in [bnd_3,bnd_4]$ and set $u_1 = (\max(1-u_2^2 - u_3^2,0))^{1/2}$. We then generate two candidates $\mathbf{u}^{\pm} = (\pm u_1 , u_2, u_3) /\norm{(u_1, u_2, u_3)}_2$ from which only one can be on the valid side of the separating plane satisfying inequality~\eqref{LowerArmLegConst}. To check, we first undo the projection and normalization by $\mathbf{b}^{\pm} = l \hspace{0.5mm} \mathbf{T}^{-1} \mathbf{u}^{\pm}$, where $l$ is a sample from the bone length distribution on $\mathbf{b}$. A sample ``$\mathbf{b}$'' is accepted only if it satisfies inequality~\eqref{LowerArmLegConst}. Note that similar bones have the same length therefore we sample their length only once for each pose. The prior model can be written as below according to a Bayesian graph on the kinematic chain:
\vspace{-1.5mm}
\begin{align}
\label{Prior}
&p(\mathbf{X}) = p(\mathbf{X}_{i \in \text{torso}}) p(\mathbf{X}_{\text{head}} | \mathbf{X}_{i \in \text{torso}}) \times \nonumber \\
& p(\mathbf{X}_{i \in \text{ l/r elbow}} | \mathbf{X}_{i \in \text{torso}}) p(\mathbf{X}_{i \in \text{ l/r wrist}} | \mathbf{X}_{i \in \text{ l/r elbow}}, \mathbf{X}_{i \in \text{torso}}) \times \nonumber \\
& p(\mathbf{X}_{i \in \text{ l/r knee}} | \mathbf{X}_{i \in \text{torso}}) p(\mathbf{X}_{i \in \text{ l/r ankle}} | \mathbf{X}_{i \in \text{ l/r knee}}, \mathbf{X}_{i \in \text{torso}}),
\vspace{-1.5mm}
\end{align}
where $p(\mathbf{X}_{i \in \text{torso}})$ is the probability of selecting a torso from the torso dictionary which we assumed is uniform. The torso joints $\mathbf{X}_{i \in \text{torso}}$ are used to determine the local coordinate system for the rest of the joints. We have removed torso joints in the equations below for notational convenience. We have:
\vspace{-1.5mm}
\begin{align}
\label{firstOrder}
&p(\mathbf{X}_i) = \frac{1}{l_{\text{bone}}^2 |\sin(\phi_i)|} p(l_{\text{bone}}) \hspace{0.5mm} p(\mathbf{\theta}_i,\mathbf{\phi}_i),
\end{align}
for $(i,\text{bone})$ being from {(l/r knee, upper leg) , (head, neck + head bone), or (l/r elbow, upper arm). The multiplier factor in~\eqref{firstOrder}, which is the inverse of Jacobian determinant for a transformation from the Cartesian to spherical coordinate system, is to ensure that the left side sums up to one if $\int_l \int_{\theta} \int_{\phi} p(l) \hspace{0.5mm} p(\mathbf{\theta},\mathbf{\phi}) d\phi \hspace{0.5mm} d\theta \hspace{0.5mm} dl  = 1$, since $dx \hspace{0.5mm} dy \hspace{0.5mm} dz = l^2 |\sin(\phi)| \hspace{0.5mm} dl \hspace{0.5mm} d\theta \hspace{0.5mm} d\phi$. For lower limbs we have:
\begin{align}
\label{LowerArmLeg}
 p(\mathbf{X}_i | \mathbf{X}_{pa(i)}) \propto p(l_{\text{bone}}) \mathbf{1}_{valid}(\mathbf{X}_i,\mathbf{X}_{pa(i)})
\end{align}
where $(i,pa(i),\text{bone})$ is from {(l/r wrist, l/r elbow, forearm) or (l/r ankle, l/r knee, lower leg), and $\mathbf{1}_{valid}(\mathbf{X}_i,\mathbf{X}_{pa(i)})$ is an indicator function that nulls the probability of configurations whose angles does not satisfy the constraints in inequalities~\eqref{LowerArmLegConst} and~\eqref{BoundingBoxConst} for $\mathbf{b} = \mathbf{X}_i - \mathbf{X}_{pa(i)}$. Conditional sampling is carried out by rejection sampling discussed in the next subsection.


\begin{figure*}[t]
\centering
\includegraphics[height=0.16\textwidth, width=0.15\textwidth]{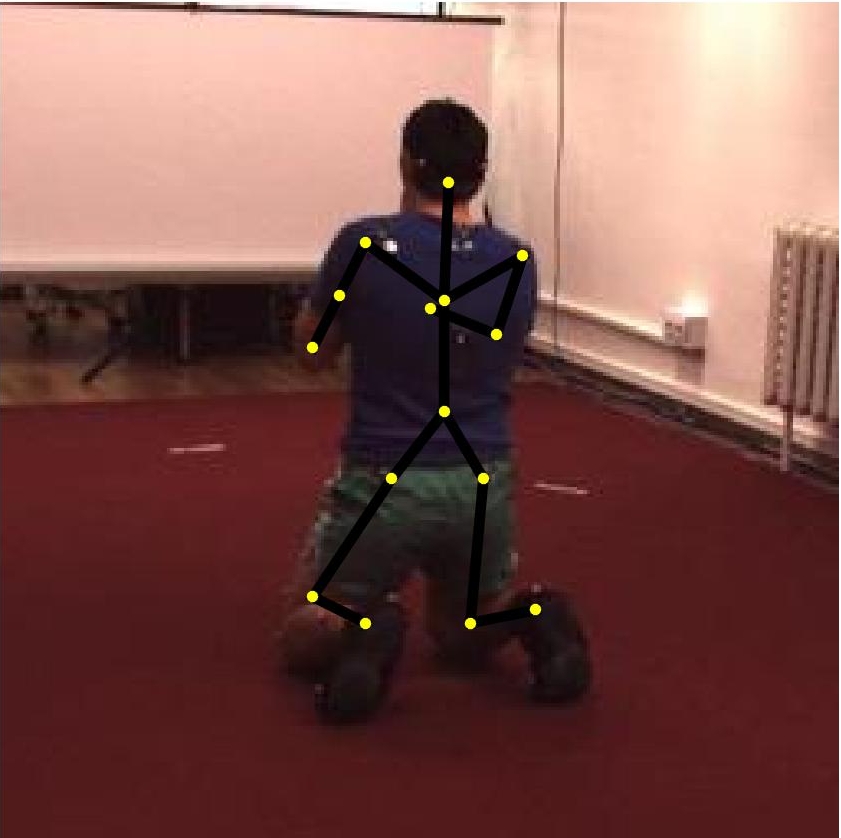}
\includegraphics[height=0.16\textwidth, width=0.15\textwidth]{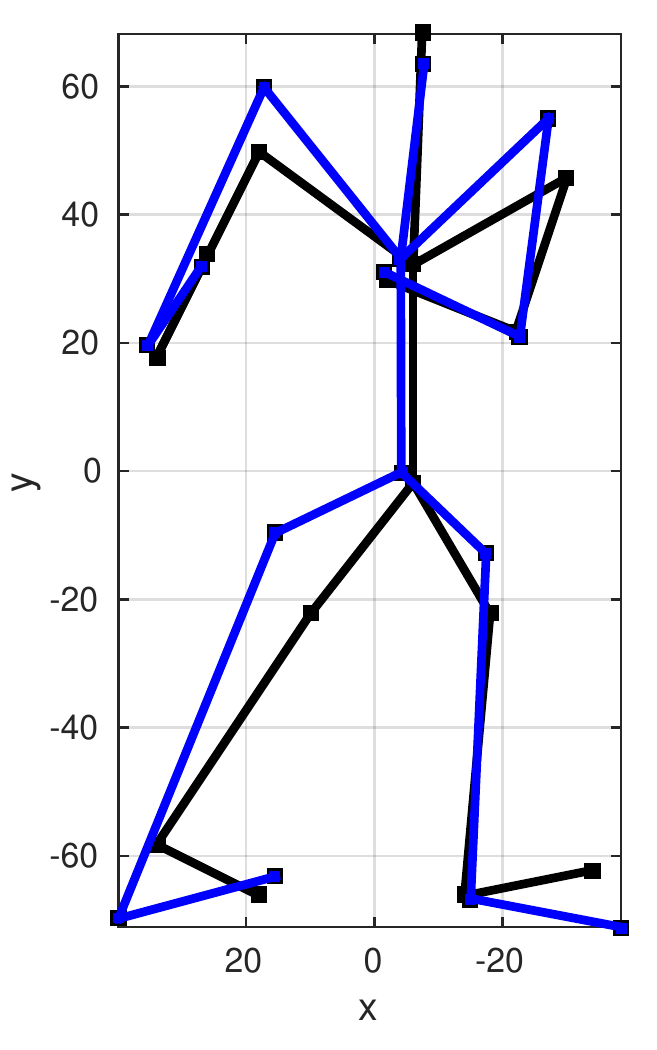}
\includegraphics[height=0.16\textwidth, width=0.15\textwidth]{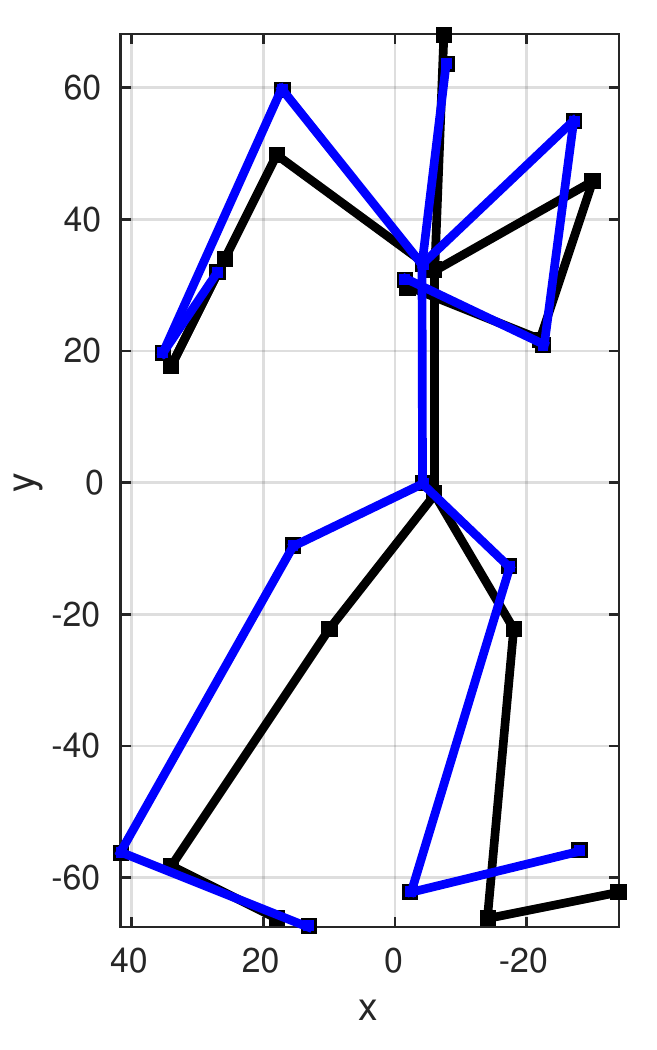}
\includegraphics[height=0.16\textwidth, width=0.15\textwidth]{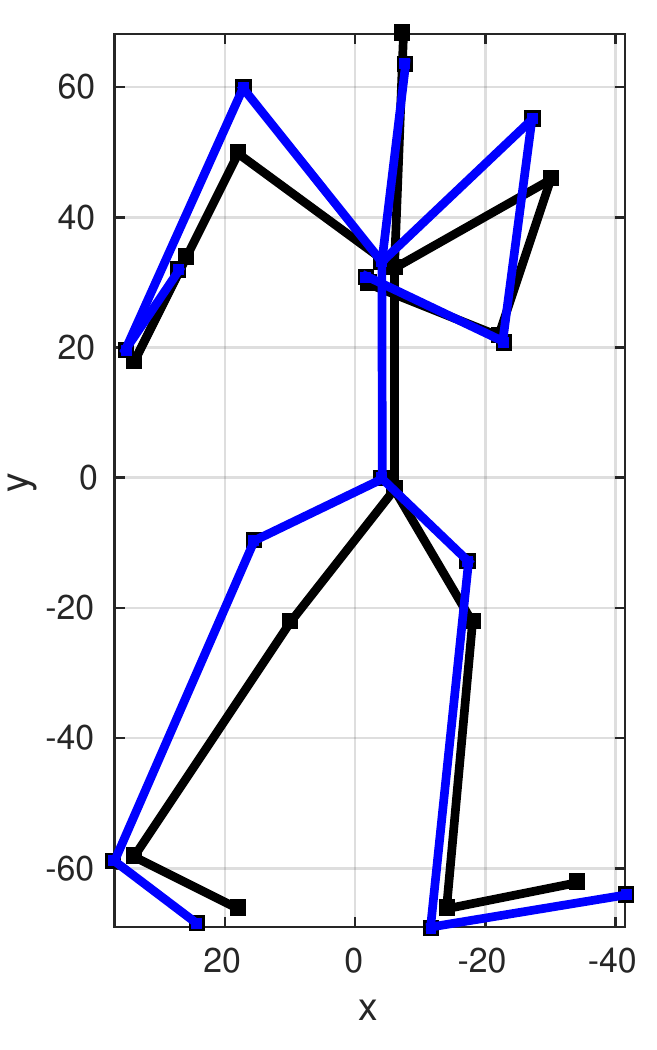}
\includegraphics[height=0.16\textwidth, width=0.15\textwidth]{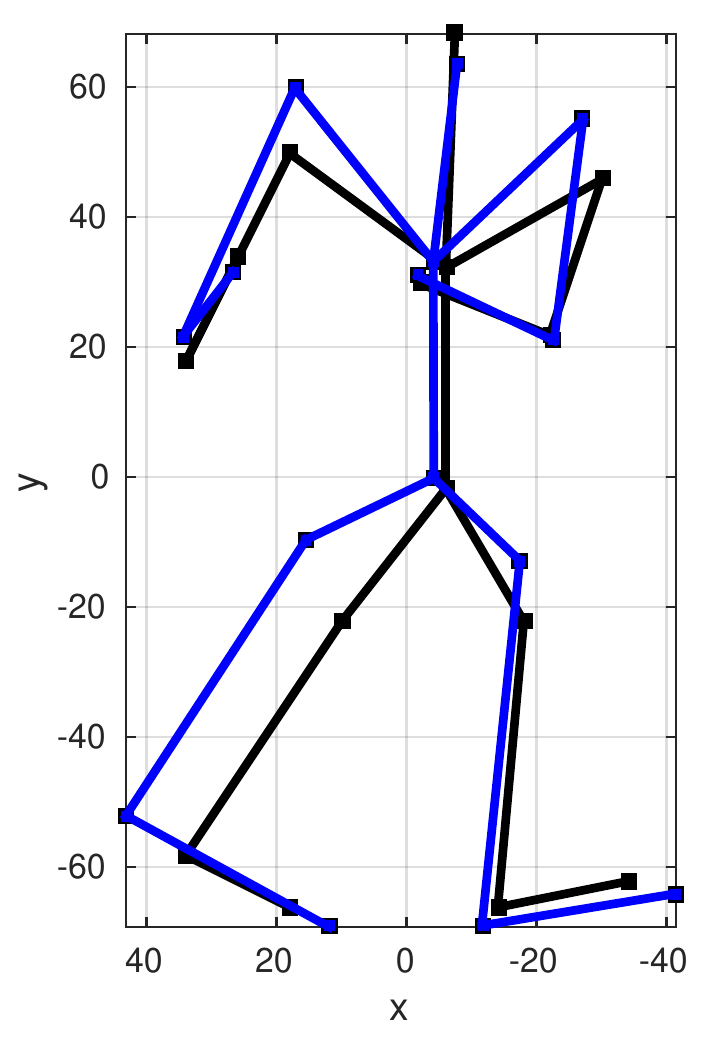}
\includegraphics[height=0.16\textwidth, width=0.15\textwidth]{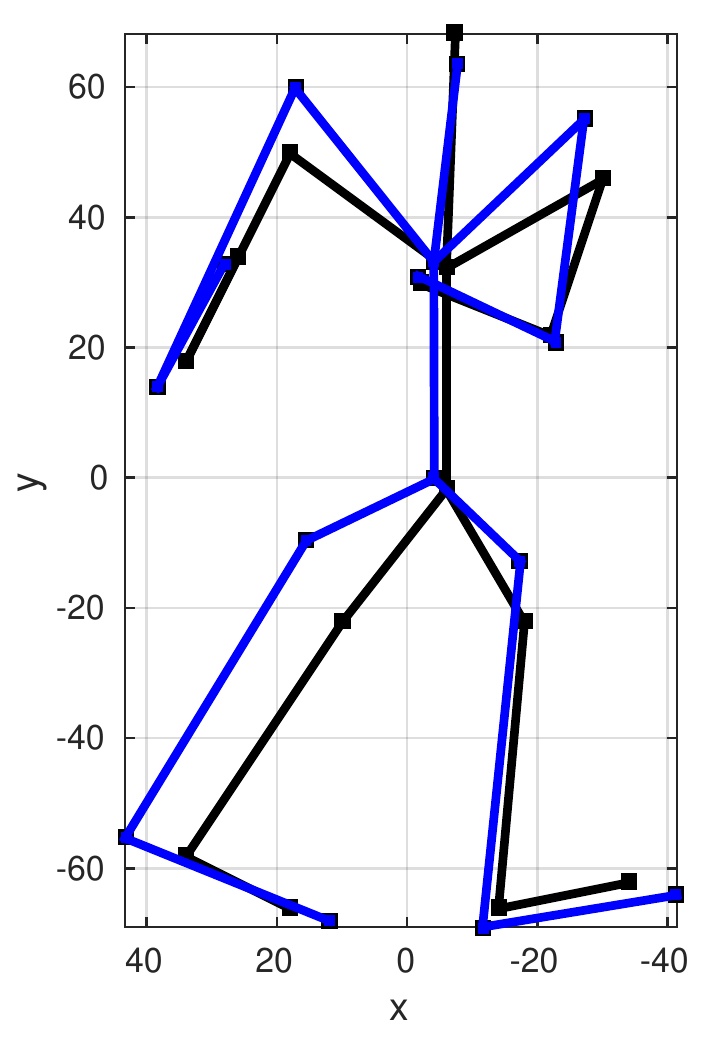} \\
\includegraphics[height=0.2\textwidth, width=0.15\textwidth]{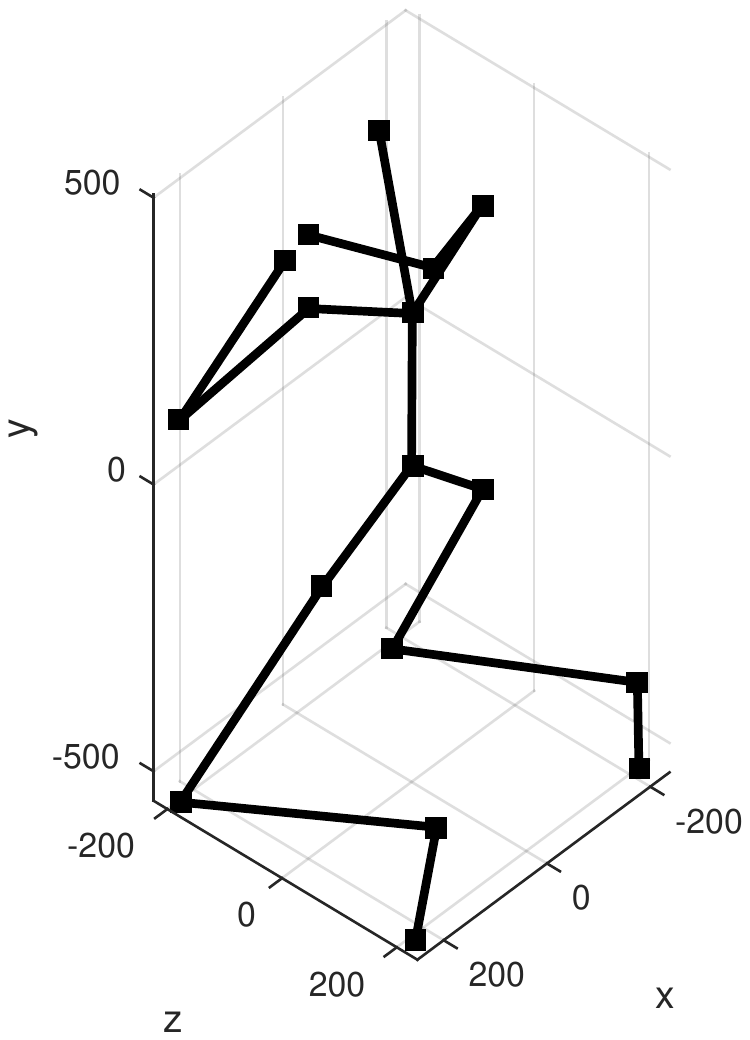} 
\includegraphics[height=0.18\textwidth, width=0.15\textwidth]{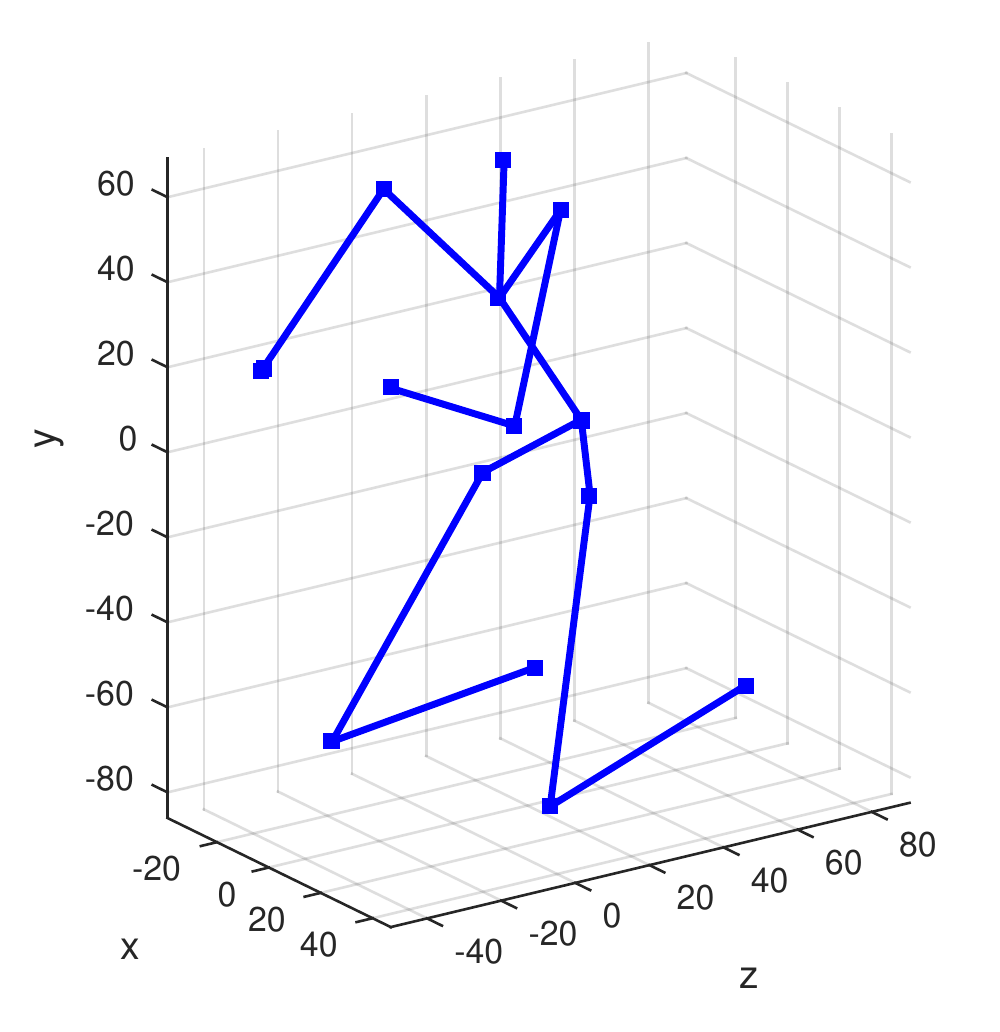}
\includegraphics[height=0.18\textwidth, width=0.15\textwidth]{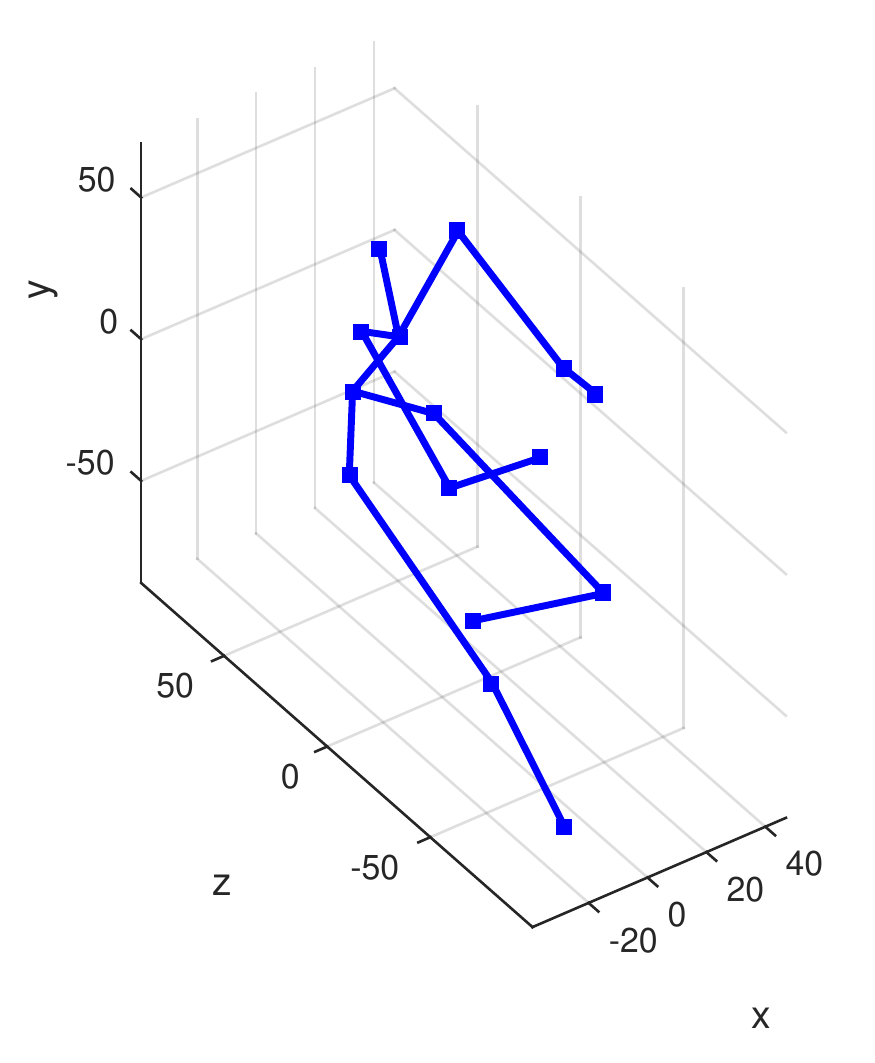}
\includegraphics[height=0.18\textwidth, width=0.15\textwidth]{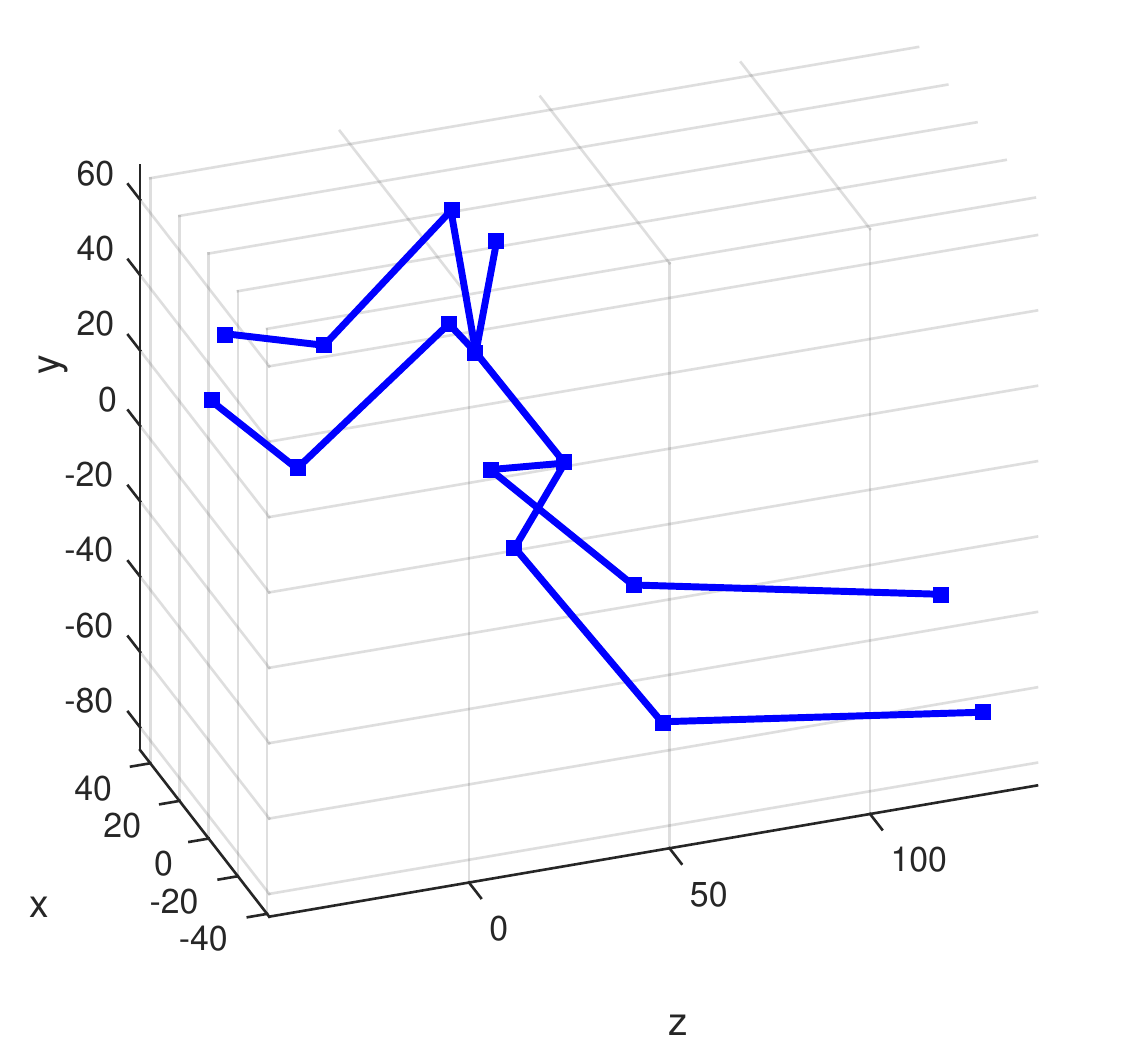}
\includegraphics[height=0.18\textwidth, width=0.15\textwidth]{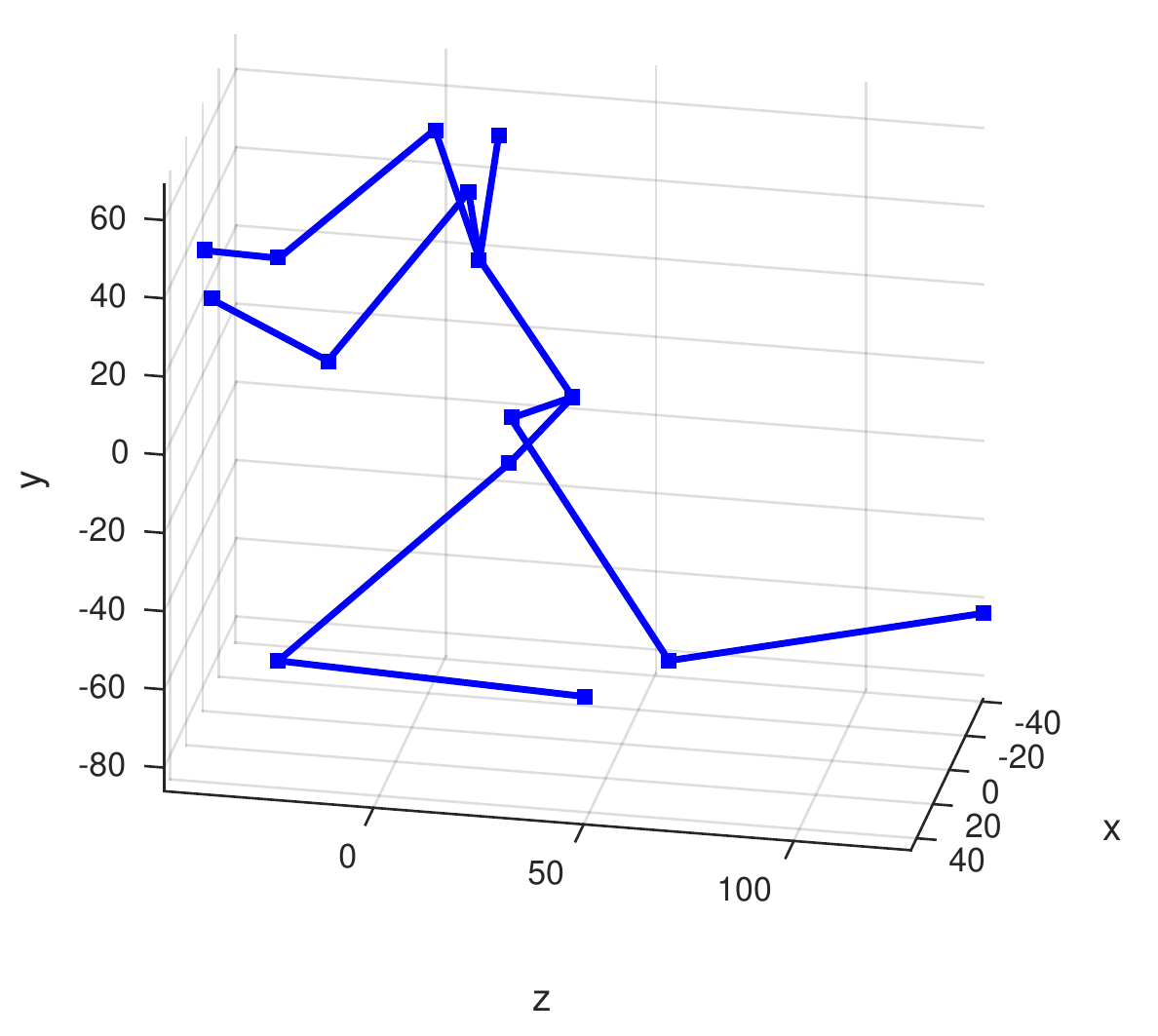}
\includegraphics[height=0.18\textwidth, width=0.15\textwidth]{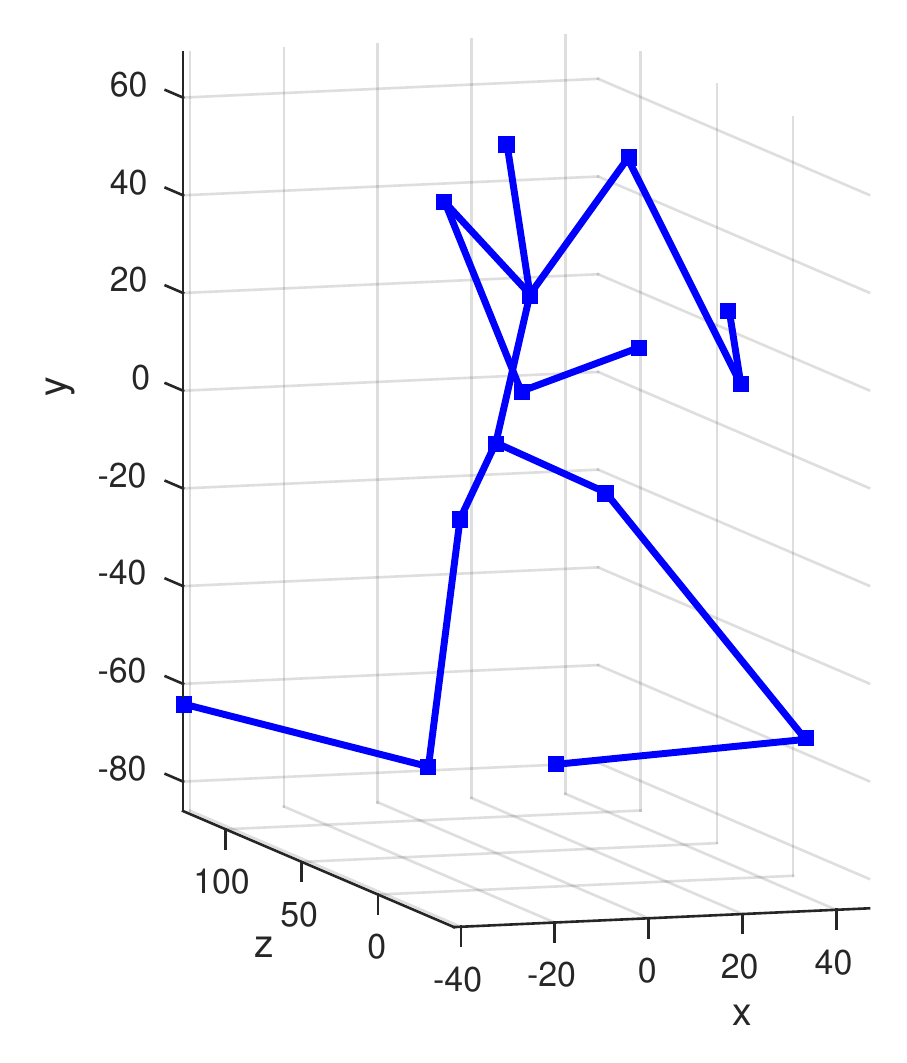}
\\
\hspace{-5.5cm} (a) \hspace{7cm} (b)
\vspace{1mm}
\caption[Diverse Hypotheses]{\footnotesize{(a): The input image and the corresponding 3D pose. (b): Generation of five diverse 3D pose hypotheses consistent with the 2D joint detections. 
}}
\label{fig:ConditionalSamples}
\vspace{-2mm}
\end{figure*}

\vspace{2mm}
\subsection{Conditional Sampling}
\label{Subsec:CondSamp}
We run a 2D joint detector on the input image $I$ and get an estimate of the 2D joint locations $\hat{\mathbf{x}}$ with confidence scores $\mathbf{\alpha}$. Then, to obtain a reasonable estimate of torso $\hat{\mathbf{X}}_{i \in \text{torso}}$ and camera parameters namely $(\hat{\mathbf{R}},\hat{\mathbf{t}},\hat{s})$, we run a 2D-to-3D pose estimator capable of handling missing joints (we modified~\cite{Akhter2015} and~\cite{ZhouUpenn2015} to handle missing joints; see equation~\eqref{Cr_missing}). Note that we are not restricted to any particular 2D/3D pose estimator and any 2D joint detector that estimates 2D joint locations $\hat{\mathbf{x}}$ and their confidence scores $\mathbf{\alpha}$ and any 2D-to-3D pose estimator can be used in the initial stage. We then assume that the estimated camera parameters and $\hat{\mathbf{X}}_{i \in \text{torso}}$ are reasonably well estimated and keep them fixed. Note that the human torso and its pose (usually vertical) does not vary much compared to the whole body pose. We do not include the estimated camera parameters and 3D torso in our formulation below for notational convenience. From the Bayes rule we have:
\begin{align}
\label{Posterior}
p(\mathbf{X} | \hat{\mathbf{x}}, \mathbf{\alpha}) \propto p(\mathbf{X}) \hspace{0.5mm} p(\hat{\mathbf{x}}, \mathbf{\alpha} | \mathbf{X}).
\vspace{-2mm}
\end{align}
We define:
\begin{align*}
p(\hat{\mathbf{x}}, \mathbf{\alpha} | \mathbf{X}) \propto
\prod_{i \in  \text{ limb } \cap S_{\text{o}}} \mathbf{1}(\norm{\hat{\mathbf{x}}_i - \hat{s} \hspace{0.5mm} \hat{\mathbf{R}}_{1:2} \hspace{0.5mm} \mathbf{X}_i + \hat{\mathbf{t}}}_{2} < \tau_i)
\vspace{-2mm}
\end{align*}
where $\mathbf{1}(.)$ is the indicator function depending on the 2D distance between detected joints and the projected 3D pose under an acceptance threshold defined by $\tau_i = 0.25 \hspace{0.5mm} \hat{s} \hspace{0.5mm} \bar{l}_{\text{limb}} / \alpha_i$, where $\bar{l}_{\text{limb}}$ is the mean limb length, $\hat{s}$ is the estimated scaling factor, $\alpha_i$ is the $i^{\text{th}}$ joint normalized confidence score, and the factor $0.25$ was chosen empirically. The likelihood function defined above accepts prior (unconditional) samples $\mathbf{X}^{(q)} \sim p(\mathbf{X})$ whose projected joints to the image coordinate system are within a distance not greater than thresholds $\tau_i$ from detected limb joints. The inverse proportion of the threshold to the confidence $\alpha_i$ allows acceptance in a larger area if the confidence score is smaller for the $i^{\text{th}}$ limb joint and therefore considering the 2D joint detection uncertainty. Note that there is no indicator function in the likelihood function for the missing limb joints which allows acceptance of all anatomically plausible samples for limb joints from $S_{\text{m}}$. Note that even though torso pose estimation is a much easier problem compared to the full body pose estimation, a poorly estimated torso, \eg due to occlusion, can adversely affect the quality of conditional 3D pose samples.


\vspace{2mm}
\subsection{Generating Diverse Hypotheses}
\label{Subsec:DivHyp}
The diversification is implemented in two stages: (I) we sampled the occupancy matrix at 15 equidistant azimuth and 15 equidistant polar angles for the upper limbs and accept the samples if the occupancy matrix had a 1 at these locations. For the lower limbs, we sampled 5 equidistant points along each $u_2$ and $u_3$ directions between $[bnd_1,bnd_2]$ and $[bnd_3,bnd_4]$, respectively. (II) To generate fewer number of pose hypothesis, we use the kmeans++ algorithm~\cite{kmenaspp2007} to cluster the posterior samples into a desired number of diverse clusters and take the nearest neighbor 3D pose sample to each centroid as one hypothesis. Kmeans++ operates the same as Kmeans clustering except that it uses a diverse initialization method to help with diversification of final clusters. Note that we cannot take the centroids as hypotheses since there is no guarantee that the mean of 3D poses is still a valid 3D pose. Figure~\ref{fig:ConditionalSamples} shows five hypotheses given the output of Hourglass 2D joint detector for the top-left image and detections shown by yellow points. In Figure~\ref{fig:ConditionalSamples}, the 2D detection of joints are shown by the black skeleton and the diversified hypotheses that are consistent with the 2D detections are shown by the blue skeletons. It can be seen that even though the 2D projection of these pose hypotheses are very similar, they are quite different in 3D. To generate the pose hypotheses in Figure~\ref{fig:ConditionalSamples}, we estimated the 3D torso and projection matrix using~\cite{Akhter2015}.
s

\section{Experimental Results}
\label{Sec:Results}

\begin{table*}[t] 
\footnotesize
\vspace{1mm}  
\begin{tabular}{l*{9}{c}} 
\toprule
{\textbf{Method}} & {\textbf{Directions}} & {\textbf{Discussion}} & {\textbf{Eating}} & {\textbf{Greeting}} & {\textbf{Phoning}} & {\textbf{Posing}} & {\textbf{Purchases}} & {\textbf{Sitting}} & {\textbf{SitDown}} \\ 
\midrule
Ours (No KM++/\cite{ZhouUpenn2015}) & \textbf{63.12} & \textbf{55.91} & \textbf{58.11} & \textbf{64.48} & \textbf{68.69} & \textbf{61.27} & \textbf{55.57} & \textbf{86.06} & \textbf{117.57} \\ 
Ours (k=20/\cite{ZhouUpenn2015}) & 77.08 & 71.15 & 75.39 & 79.01 & 84.68 & 74.90 & 72.37 & 102.17 & 131.46 \\ 
Ours (k=5/\cite{ZhouUpenn2015}) & 82.86 & 77.52 & 81.60 & 85.20 & 90.93 & 80.46 & 78.75 & 109.27 & 138.71 \\ 
Zhou et al.~\cite{ZhouUpenn2015} & 80.51 & 74.56 & 73.95 & 85.43 & 88.96 & 82.02 & 76.21 & 107.43 & 146.47 \\
\hdashline
Ours (k=5/\cite{Akhter2015}) & \textbf{105.14} & \textbf{100.28} & \textbf{107.75} & \textbf{106.88} & \textbf{111.44} & \textbf{105.74} & \textbf{101.18} & \textbf{124.87} & \textbf{147.48} \\ 
Akhter\&Black~\cite{Akhter2015} & 133.80 & 128.03 & 124.47 & 133.47 & 133.93 & 136.63 & 128.30 & 133.61 & 162.01 \\ 
\hdashline
Chen et al.~\cite{Deep3DPose} & 145.37 & 139.11 & 140.24 & 149.13 & 149.61 & 154.30 & 147.04 & 161.49 & 200.06 \\ 
\bottomrule
\end{tabular}%
\vspace{2.5mm}    
\begin{tabular}{l*{8}{c}} 
 & {\textbf{Smoking}} & {\textbf{TakingPhoto}} & {\textbf{Waiting}} & {\textbf{Walking}} & {\textbf{WalkingDog}} & {\textbf{WalkTogether}} & {\textbf{Average}} & \\ 
\midrule
Ours (No KM++/\cite{ZhouUpenn2015}) & \textbf{71.02} & \textbf{71.21} & \textbf{66.29} & \textbf{57.07} & \textbf{62.50} & \textbf{61.02} & \textbf{67.99} \\ 
Ours (k=20/\cite{ZhouUpenn2015}) & 85.90 & 84.49 & 80.41 & 71.57 & 78.41 & 74.92 & 82.93 \\ 
Ours (k=5/\cite{ZhouUpenn2015}) & 91.79 & 90.06 & 86.43 & 77.93 & 85.45 & 81.49 & 89.23 \\ 
Zhou et al.~\cite{ZhouUpenn2015} & 90.61 & 93.43 & 85.71 & 80.03 & 90.89 & 85.73 & 89.46 \\ 
\hdashline  
Ours (k=5/\cite{Akhter2015}) & \textbf{113.61} & \textbf{105.58} & \textbf{105.80} & \textbf{100.28} & \textbf{106.25} & \textbf{104.63} & \textbf{109.79} \\ 
Akhter\&Black~\cite{Akhter2015} & 135.75 & 132.92 & 133.93 & 133.84 & 131.77 & 134.80 & 134.48 \\ 
\hdashline
Chen et al.~\cite{Deep3DPose} & 152.37 & 159.18 & 152.67 & 148.20 & 156.10 & 147.71 & 153.51 \\ 
\bottomrule
\end{tabular}   \\
\caption{\footnotesize{Quantitative comparison on the Human3.6M dataset evaluated in 3D by mean per joint error (mm) for all actions and subjects whose ground-truth 3D poses were provided.}}
\label{tab:MeanError}
\vspace{-3mm}
\end{table*}

We empirically evaluated the proposed ``multi-pose hypotheses'' approach on the recently published Human3.6M dataset~\cite{h36m_pami}. 
For evaluation, we used images from all 4 cameras and all 15 actions associated with 7 subjects for whom ground-truth 3D poses were provided namely subjects S1, S5, S6, S7, S8, S9, and S11. 
The original videos (50 fps) were downsampled (in order to reduce the correlation of consecutive frames) to built a dataset of 26385 images. For further evaluation, we also built two rotation datasets by rotating H36M images by $30$ and $60$ degrees. 
We evaluated the performance by the mean per joint error (millimeter) in 3D by comparing the reconstructed pose hypotheses against the ground truth. The error was calculated up to a similarity transformation obtained by Procrustes alignment. The results are summarized in Table~\ref{tab:MeanError} for various methods and actions. For a fair comparison, the limb length of the reconstructed poses from all methods were scaled to match the limb length of the ground-truth pose. The bone length matching obviously lowers the mean joint errors but makes no difference in our comparisons. One can see that the best (lowest Euclidean distance from the ground-truth pose) out of only 5 generated hypotheses by using~\cite{Akhter2015} as baseline for 3D torso and projection matrix estimation is considerably better than the single 3D pose output by~\cite{Akhter2015} for all actions. We also used the 2D-to-3D pose estimator by Zhou et al.~\cite{ZhouUpenn2015} with convex-relaxation as baseline and observed considerable improvement compared to~\cite{Akhter2015} in both 3D pose and projection matrix estimation. Using~\cite{ZhouUpenn2015} as baseline to estimate the 3D torso and projection matrix we generated multiple 3D pose hypotheses. Since the accuracy of~\cite{ZhouUpenn2015} is already high, the best out of 5 pose hypotheses cannot significantly lower the average joint distance from the single 3D pose output by~\cite{ZhouUpenn2015}. However, by increasing the number of hypotheses we started to observe improvement. Table~\ref{tab:MeanError} also includes the best hypothesis out of conditional samples from only the first diversification stage \ie, by diversifying conditional samples and using no kmeans++ clustering (shown by No KM++), using~\cite{ZhouUpenn2015} as base. This achieves the lowest joint error in comparison to other baselines. The pose hypotheses can be generated very quickly ($<$ 2 seconds) in Matlab on an Intel i7-4790K processor.


\begin{figure*}[t]
\centering
\includegraphics[width=0.32\textwidth]
{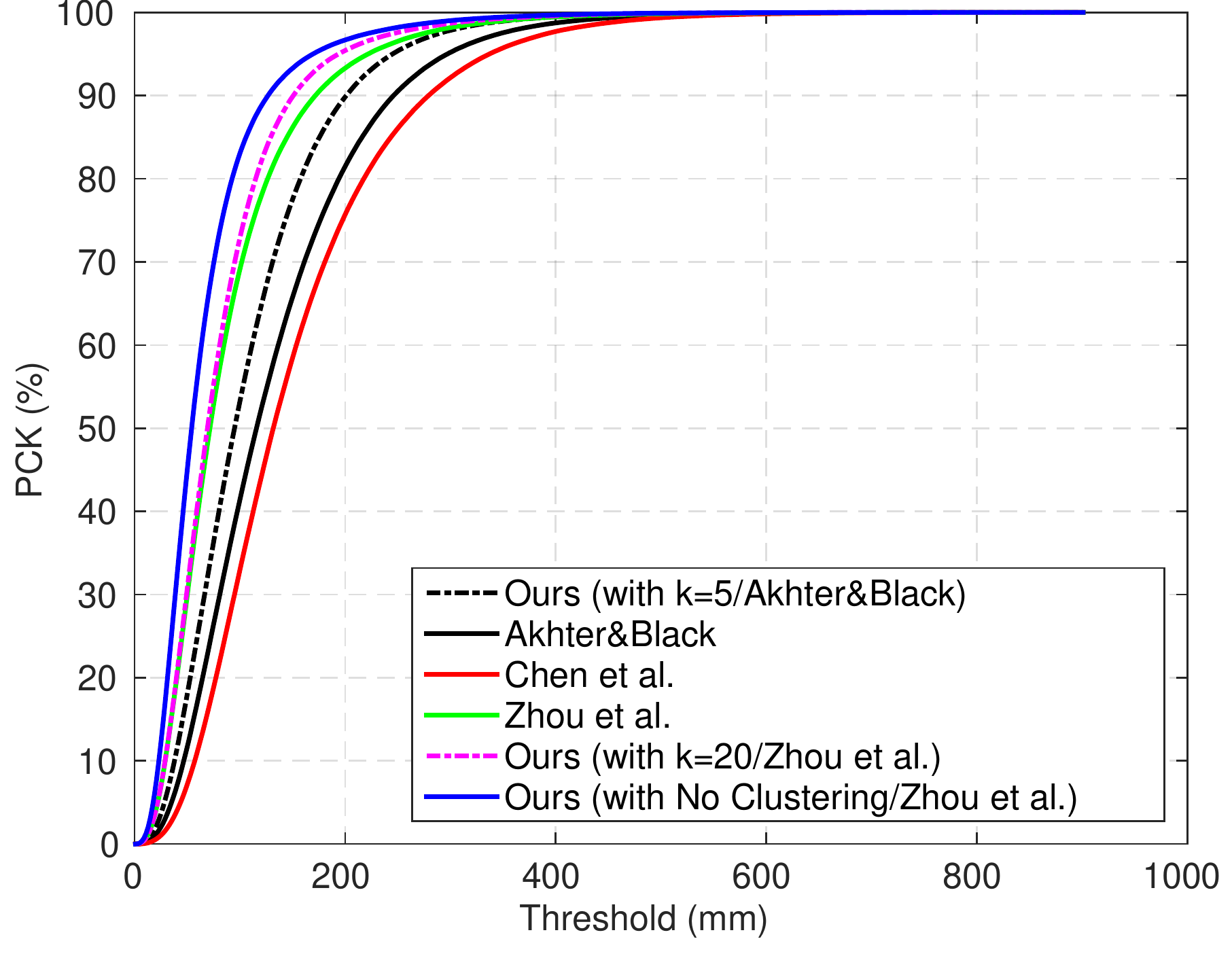} 
\includegraphics[width=0.32\textwidth]{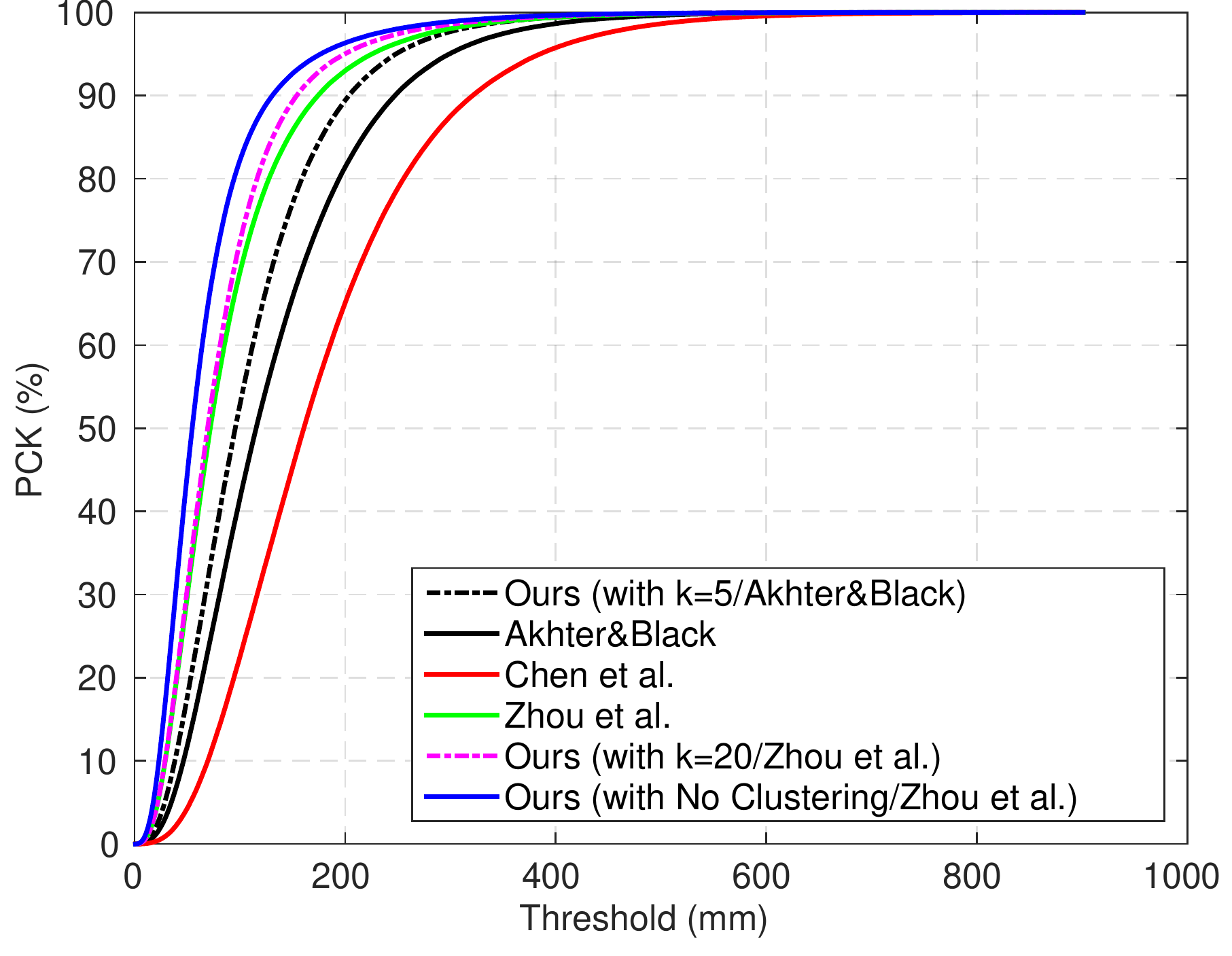} 
\includegraphics[width=0.32\textwidth]{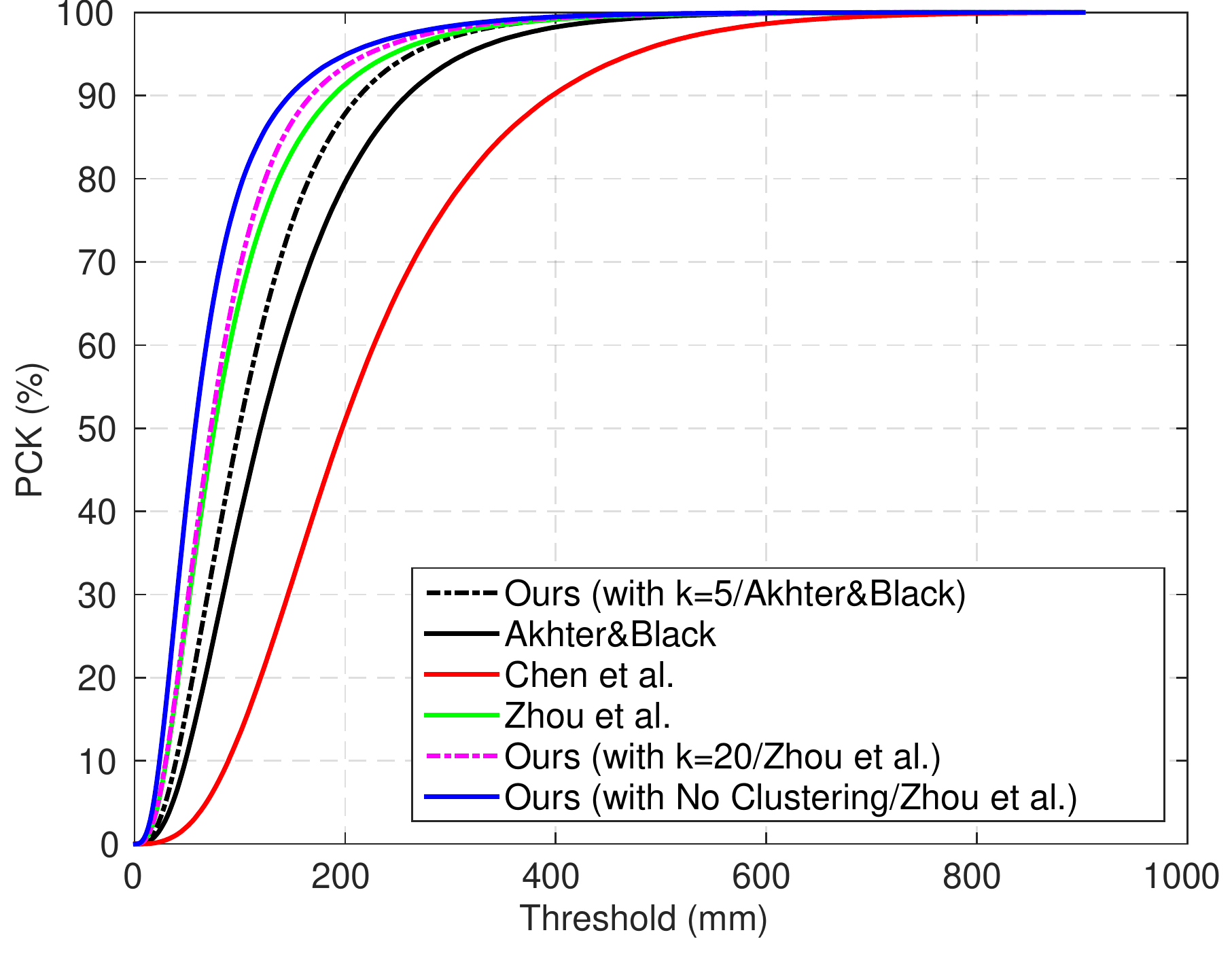} 
\vspace{-2mm}
\caption[PCK]{\footnotesize{PCK curves for the H36M dataset (original), H36M rotated by 30 and 60 degrees respectively from left to right. The y-axis is the percentage of correctly detected joints in 3D for a given distance threshold in millimeter (x-axis).}}
\label{fig:PCK}
\vspace{-2mm}
\end{figure*}

\begin{table*}[h]
\footnotesize
\vspace{1mm}  
\begin{tabular}{l*{9}{c}} 
\toprule
{\textbf{Method}} & {\textbf{Directions}} & {\textbf{Discussion}} & {\textbf{Eating}} & {\textbf{Greeting}} & {\textbf{Phoning}} & {\textbf{Posing}} & {\textbf{Purchases}} & {\textbf{Sitting}} & {\textbf{SitDown}} \\ 
\midrule
Ours (k=5/\cite{Akhter2015}) & \textbf{98.44} & \textbf{93.70} & \textbf{102.62} & \textbf{97.50} & \textbf{96.29} & \textbf{98.90} & \textbf{93.32} & \textbf{105.51} & \textbf{110.07} \\  
Akhter\&Black~\cite{Akhter2015} & 118.02 & 112.55 & 111.27 & 117.46 & 111.77 & 122.27 & 112.23 & 107.27 & 126.95 \\
\hdashline
Ours (k=5/\cite{Akhter2015}) & \textbf{108.60} & \textbf{105.85} & \textbf{105.63} & \textbf{109.01} & \textbf{105.47} & \textbf{109.93} & \textbf{102.01} & \textbf{111.25} & \textbf{119.57} \\ 
Akhter\&Black~\cite{Akhter2015} & 153.80 & 149.14 & 135.44 & 155.06 & 139.62 & 156.46 & 149.05 & 126.33 & 141.89 \\
\hdashline
Ours (k=5/\cite{Akhter2015}) & \textbf{125.03} & \textbf{121.77} & \textbf{115.13} & \textbf{124.11} & \textbf{116.92} & \textbf{123.75} & \textbf{116.42} & \textbf{119.63} & \textbf{130.81} \\ 
Akhter\&Black~\cite{Akhter2015} & 185.57 & 180.43 & 158.55 & 185.65 & 162.39 & 185.78 & 178.81 & 145.15 & 155.29 \\ 
\bottomrule
\end{tabular}%
\vspace{1.5mm}    
\begin{tabular}{l*{8}{c}} 
 & {\textbf{Smoking}} & {\textbf{TakingPhoto}} & {\textbf{Waiting}} & {\textbf{Walking}} & {\textbf{WalkingDog}} & {\textbf{WalkTogether}} & {\textbf{Average}} & {\textbf{Average Diff.}} \\ 
\midrule
Ours (k=5/\cite{Akhter2015}) & \textbf{97.53} & \textbf{97.63} & \textbf{99.43} & \textbf{90.23} & \textbf{97.27} & \textbf{95.21} & \textbf{98.24} & \\ 
Akhter\&Black~\cite{Akhter2015} & 113.22 & 120.61 & 119.97 & 115.81 & 116.60 & 115.62 & 116.11 & $\boxed{17.87}$ \\ 
\hdashline  
Ours (k=5/\cite{Akhter2015}) & \textbf{107.76} & \textbf{107.05} & \textbf{111.34} & \textbf{108.38} & \textbf{106.96} & \textbf{110.28} & \textbf{108.61} &  \\
Akhter\&Black~\cite{Akhter2015} & 142.98 & 152.65 & 155.27 & 155.18 & 151.88 & 155.00 & 147.98 & $\boxed{39.37}$ \\
\hdashline  
Ours (k=5/\cite{Akhter2015}) & \textbf{120.60} & \textbf{118.38} & \textbf{127.13} & \textbf{125.89} & \textbf{121.61} & \textbf{127.62} & \textbf{122.32} & \\
Akhter\&Black~\cite{Akhter2015} & 165.47 & 177.44 & 186.20 & 189.66 & 183.01 & 186.25 & 175.04 & $\boxed{\textbf{52.72}}$ \\ 
\bottomrule
\end{tabular}   \\
\caption{\footnotesize{Quantitative comparison on the Human3.6M dataset when 0 (top pair), 1 (middle pair), and 2 (bottom pair) limb joints are missing.}}
\label{tab:MissingJoints}
\vspace{-2mm}
\end{table*}

We also used Deep3D of Chen et al.~\cite{Deep3DPose} as another baseline. The Deep3D~\cite{Deep3DPose} is a 3D pose estimator that directly regresses to the 3D joint locations directly from a monocular RGB input image. Deep3D had the highest mean joint errors as shown in Table~\ref{tab:MeanError}. We also observed that the pre-trained Deep3D is very sensitive to image rotation and usually outputs an anatomically implausible 3D pose if the input image is rotated. But other 2D-to-3D pose estimation baselines which decouple the projection matrix and the 3D pose are quite robust to rotation of the input image. Figure~\ref{fig:PCK} shows the Percentage of Correct Keypoints (PCK) versus an acceptance distance threshold in millimeter for various baselines and H36M dataset variations namely the original H36M and 30/60 degree rotations. One can see that the PCK of Deep3D drops drastically by rotating the input image. This is partly due to insufficient number of tilted samples in the training set (H36M plus synthetic images). One of the main problems of purely discriminative approaches such as~\cite{Deep3DPose} is their extreme sensitivity to data manipulation. On the other hand, humans can learn from a few examples and still not suppress the rarely seen cases compared to the frequently seen ones. 

In a realistic scenario with occlusion, the location of some 2D joints cannot be accurately detected. The added uncertainty caused by occlusion makes one expect a larger average estimation error for the estimated 3D pose from a single-output pose estimator compared to the best 3D pose hypothesis. To test this, we ran experiments with different number of missing joints (0, 1 and 2) selected randomly from the limb joints including l/r elbow, l/r wrist, l/r knee, and l/r ankle. Table~\ref{tab:MissingJoints} shows the mean per joint errors for the 3D pose estimated by the modified version of Akhter\&Black~\cite{Akhter2015} that can handle missing joints compared to the best out of five hypotheses generated by our method when 0, 1, and 2 limb joints are missing. In this test, we used the ground-truth 2D location of the joints and randomly selected the missing joints. 
One can see that by increasing the number of missing joints the performance gap between the estimated 3D pose and the best 3D pose hypothesis increases. This underscores the importance of having multiple hypothesis for more realistic scenarios.

\vspace{-2mm}
\section{Conclusion}
\label{Sec:Conclusion}
\vspace{-1mm}
There usually exist multiple 3D poses consistent with the 2D location of joints because of losing the depth information in monocular images. The uncertainty in 3D pose estimation increases in the presence of occlusion and imperfect 2D detection of joints. In this paper, we proposed a way to generate multiple valid and diverse 3D pose hypotheses consistent with the 2D joint detections. These pose hypotheses can be ranked later by more detailed investigation of the image beyond the 2D joint locations or based on some contextual information. To generate these pose hypotheses we used a novel unbiased generative model that only enforces pose-conditioned anatomical constraints on the joint-angle limits and limb length ratios. This was motivated by the pose-conditioned joint limits from~\cite{Akhter2015} after identifying bias in typical MoCap datasets. Our compositional generative model uniformly spans the full variability of human 3D pose which helps in generating more diverse hypotheses. We performed empirical evaluation on the H36M dataset and achieved lower mean joint errors for the best pose hypothesis compared to the estimated pose by other recent baselines. The 3D pose output by the baseline methods could also be included as one hypothesis but to investigate our hypothesis generation approach we did not do so in the experimental results. Our experiments show the importance of having multiple 3D pose hypotheses given only the 2D location of joints especially when some of the joints are missing. We hope our idea of generating multiple pose hypotheses inspire a new line of future work in 3D pose estimation considering various ambiguity sources.

{\small
\bibliographystyle{ieee}
\bibliography{egbib}

\begin{thebibliography}{10}\itemsep=-1pt

\bibitem{Akhter2015}
I.~Akhter and M.~J. Black.
\newblock Pose-conditioned joint angle limits for {3D} human pose
  reconstruction.
\newblock In {\em CVPR}, pages 1446--1455, June 2015.

\bibitem{amin13bmvc}
S.~Amin, M.~Andriluka, M.~Rohrbach, and B.~Schiele.
\newblock Multi-view pictorial structures for 3d human pose estimation.
\newblock In {\em British Machine Vision Conference (BMVC)}, September 2013.

\bibitem{kmenaspp2007}
D.~Arthur and S.~Vassilvitskii.
\newblock K-means++: The advantages of careful seeding.
\newblock In {\em Proceedings of the Eighteenth Annual ACM-SIAM Symposium on
  Discrete Algorithms}, SODA '07, pages 1027--1035, Philadelphia, PA, USA,
  2007. Society for Industrial and Applied Mathematics.

\bibitem{belagiannis20143d}
V.~Belagiannis, S.~Amin, M.~Andriluka, B.~Schiele, N.~Navab, and S.~Ilic.
\newblock 3d pictorial structures for multiple human pose estimation.
\newblock In {\em CVPR}, 2014.

\bibitem{belagiannis20163Dpami}
V.~Belagiannis, S.~Amin, M.~Andriluka, B.~Schiele, N.~Navab, and S.~Ilic.
\newblock 3d pictorial structures revisited: Multiple human pose estimation.
\newblock {\em IEEE Transactions on Pattern Analysis and Machine Intelligence},
  38(10):1929--1942, 2016.

\bibitem{bulat2016human}
A.~Bulat and G.~Tzimiropoulos.
\newblock Human pose estimation via convolutional part heatmap regression.
\newblock In {\em ECCV}, 2016.

\bibitem{BureniusSC13}
M.~Burenius, J.~Sullivan, and S.~Carlsson.
\newblock 3d pictorial structures for multiple view articulated pose
  estimation.
\newblock In {\em CVPR}, pages 3618--3625, 2013.

\bibitem{Deep3DPose}
W.~Chen, H.~Wang, Y.~Li, H.~Su, Z.~Wang, C.~Tu, D.~Lischinski, D.~Cohen-Or, and
  B.~Chen.
\newblock Synthesizing training images for boosting human 3d pose estimation.
\newblock In {\em 3D Vision (3DV)}, 2016.

\bibitem{chen2014articulated}
X.~Chen and A.~L. Yuille.
\newblock Articulated pose estimation by a graphical model with image dependent
  pairwise relations.
\newblock In {\em Advances in Neural Information Processing Systems}, pages
  1736--1744, 2014.

\bibitem{chu2016structured}
X.~Chu, W.~Ouyang, H.~Li, and X.~Wang.
\newblock Structured feature learning for pose estimation.
\newblock In {\em CVPR}, 2016.

\bibitem{CMUDataset}
P.~Doe.
\newblock Cmu human motion capture database. availabel online at:, 2003.

\bibitem{Eichner12}
M.~Eichner, M.~Marin-Jimenez, A.~Zisserman, and V.~Ferrari.
\newblock 2d articulated human pose estimation and retrieval in (almost)
  unconstrained still images.
\newblock {\em International Journal of Computer Vision}, 99:190--214, 2012.

\bibitem{Everingham15}
M.~Everingham, S.~M.~A. Eslami, L.~Van~Gool, C.~K.~I. Williams, J.~Winn, and
  A.~Zisserman.
\newblock The pascal visual object classes challenge: A retrospective.
\newblock {\em International Journal of Computer Vision}, 111(1):98--136, Jan.
  2015.

\bibitem{insafutdinov16ariv}
E.~Insafutdinov, L.~Pishchulin, B.~Andres, M.~Andriluka, and B.~Schiele.
\newblock Deepercut: A deeper, stronger, and faster multi-person pose
  estimation model.
\newblock In {\em ECCV}, May 2016.

\bibitem{h36m_pami}
C.~Ionescu, D.~Papava, V.~Olaru, and C.~Sminchisescu.
\newblock Human3.6m: Large scale datasets and predictive methods for 3d human
  sensing in natural environments.
\newblock {\em IEEE Transactions on Pattern Analysis and Machine Intelligence},
  36(7):1325--1339, jul 2014.

\bibitem{Lee2004}
M.~W. Lee and I.~Cohen.
\newblock Proposal maps driven mcmc for estimating human body pose in static
  images.
\newblock In {\em CVPR}, 2004.

\bibitem{Lehrmann2013}
A.~M. Lehrmann, P.~V. Gehler, and S.~Nowozin.
\newblock A non-parametric bayesian network prior of human pose.
\newblock In {\em CVPR}, pages 1281--1288, 2013.

\bibitem{Mallat93}
S.~G. Mallat and Z.~Zhang.
\newblock Matching pursuits with time-frequency dictionaries.
\newblock {\em IEEE Transactions on Signal Processing}, pages 3397--3415, Dec.
  1993.

\bibitem{Newell_2016}
A.~Newell, K.~Yang, and J.~Deng.
\newblock Stacked hourglass networks for human pose estimation.
\newblock In {\em ECCV}, May 2016.

\bibitem{Park2011}
D.~Park and D.~Ramanan.
\newblock N-best maximal decoders for part models.
\newblock In {\em ICCV}, 2011.

\bibitem{pishchulin16cvpr}
L.~Pishchulin, E.~Insafutdinov, S.~Tang, B.~Andres, M.~Andriluka, P.~Gehler,
  and B.~Schiele.
\newblock Deepcut: Joint subset partition and labeling for multi person pose
  estimation.
\newblock In {\em CVPR}, June 2016.

\bibitem{Moll2011}
G.~Pons-Moll, A.~Baak, J.~Gall, L.~Leal-Taixé, M.~Müller, H.-P. Seidel, and
  B.~Rosenhahn.
\newblock Outdoor human motion capture using inverse kinematics and von
  mises-fisher sampling.
\newblock In {\em ICCV}, 2011.

\bibitem{PonsMoll_CVPR2014}
G.~Pons-Moll, D.~J. Fleet, and B.~Rosenhahn.
\newblock Posebits for monocular human pose estimation.
\newblock In {\em CVPR}, pages 2345--2352, June 2014.

\bibitem{Ramakrishna_2012}
V.~Ramakrishna, T.~Kanade, and Y.~Sheikh.
\newblock Reconstructing 3d human pose from 2d image landmarks.
\newblock In {\em ECCV}, 2012.

\bibitem{rogez2016mocap}
G.~Rogez and C.~Schmid.
\newblock Mocap-guided data augmentation for 3d pose estimation in the wild.
\newblock 2016.

\bibitem{Rogez2017}
G.~Rogez, P.~Weinzaepfel, and C.~Schmid.
\newblock Lcr-net: Localization-classification-regression for human pose.
\newblock In {\em CVPR}, 2017.

\bibitem{Sapp2013}
B.~Sapp and B.~Taskar.
\newblock Modec: Multimodal decomposable models for human pose estimation.
\newblock In {\em CVPR}, pages 3674--3681, 2013.

\bibitem{HumanEva10}
L.~Sigal, A.~Balan, and M.~J. Black.
\newblock Humaneva: Synchronized video and motion capture dataset and baseline
  algorithm for evaluation of articulated human motion.
\newblock {\em International Journal of Computer Vision}, 87:4--27, 2010.

\bibitem{GMM_2004}
L.~Sigal, S.~Bhatia, S.~Roth, M.~J. Black, and M.~Isard.
\newblock Tracking loose-limbed people.
\newblock In {\em CVPR}, June 2004.

\bibitem{Sigal_IJCV_11}
L.~Sigal, M.~Isard, H.~Haussecker, and M.~J. Black.
\newblock Loose-limbed people: Estimating {3D} human pose and motion using
  non-parametric belief propagation.
\newblock {\em International Journal of Computer Vision}, 98(1):15--48, May
  2011.

\bibitem{Simo_cvpr2013}
E.~Simo-Serra, A.~Quattoni, C.~Torras, and F.~Moreno-Noguer.
\newblock A joint model for 2d and 3d pose estimation from a single image.
\newblock In {\em CVPR}, pages 3634--3641, 2013.

\bibitem{SimoSerraCVPR2012}
E.~Simo-Serra, A.~Ramisa, G.~Aleny\`a, C.~Torras, and F.~Moreno-Noguer.
\newblock Single image 3d human pose estimation from noisy observations.
\newblock In {\em CVPR}, 2012.

\bibitem{Sminchisescu2003}
C.~Sminchisescu and B.~Triggs.
\newblock Kinematic jump processes for monocular 3d human tracking.
\newblock In {\em CVPR}, 2003.

\bibitem{tompson2014joint}
J.~J. Tompson, A.~Jain, Y.~LeCun, and C.~Bregler.
\newblock Joint training of a convolutional network and a graphical model for
  human pose estimation.
\newblock In {\em Advances in neural information processing systems}, pages
  1799--1807, 2014.

\bibitem{toshev2014deeppose}
A.~Toshev and C.~Szegedy.
\newblock Deeppose: Human pose estimation via deep neural networks.
\newblock In {\em CVPR}, pages 1653--1660, 2014.

\bibitem{tSNE_2008}
L.~van~der Maaten and G.~E. Hinton.
\newblock Visualizing high-dimensional data using t-sne.
\newblock {\em Journal of Machine Learning Research}, 9:2579--2605, 2008.

\bibitem{Chunyu_2014}
C.~Wang, Y.~Wang, Z.~Lin, A.~L. Yuille, and W.~Gao.
\newblock Robust estimation of 3d human poses from a single image.
\newblock In {\em CVPR}, 2014.

\bibitem{Wei_2016}
S.-E. Wei, V.~Ramakrishna, T.~Kanade, and Y.~Sheikh.
\newblock Convolutional pose machines.
\newblock In {\em CVPR}, June 2016.

\bibitem{chu2016crf}
H.~L. X.~W. Xiao~Chu, Wanli~Ouyang.
\newblock Crf-cnn: Modeling structured information in human pose estimation.
\newblock In {\em NIPS}, 2016.

\bibitem{yang2016end}
W.~Yang, W.~Ouyang, H.~Li, and X.~Wang.
\newblock End-to-end learning of deformable mixture of parts and deep
  convolutional neural networks for human pose estimation.
\newblock In {\em CVPR}, 2016.

\bibitem{Yang2011}
Y.~Yang and D.~Ramanan.
\newblock Articulated pose estimation with flexible mixtures-of-parts.
\newblock In {\em CVPR}, 2011.

\bibitem{ZhouUpenn2015}
X.~Zhou, S.~Leonardos, X.~Hu, and K.~Daniilidis.
\newblock 3d shape estimation from 2d landmarks: A convex relaxation approach.
\newblock In {\em CVPR}, pages 4447--4455, June 2015.

\bibitem{Zhou_2016}
X.~Zhou, M.~Zhu, S.~Leonardos, K.~G. Derpanis, and K.~Daniilidis.
\newblock Sparseness meets deepness: 3d human pose estimation from monocular
  video.
\newblock In {\em CVPR}, June 2016.

\end{thebibliography}
}

\end{document}